\documentclass{article}

\usepackage{authblk}
\usepackage[utf8]{inputenc}
\usepackage{main} % ACL style file
\usepackage{microtype}
\usepackage{times}
\usepackage{latexsym}
\usepackage{algorithm}
\usepackage{algorithmic}
\usepackage{graphicx}
\usepackage{xcolor}
\definecolor{deepgreen}{RGB}{34,139,34}
\usepackage{amsmath}
\usepackage{amssymb}
\usepackage{amsfonts}
\usepackage{mathtools}
\usepackage{bm}

\usepackage{graphicx}
\usepackage{subcaption}
\usepackage{float}
\usepackage{wrapfig}

\usepackage{booktabs}
\usepackage{array}
\usepackage{multirow}
\usepackage{multicol}
\usepackage{makecell}
\usepackage{longtable}
\usepackage{tabularx}
\usepackage{colortbl}
\usepackage{arydshln}

\usepackage{color}
\usepackage{xcolor}
\definecolor{mydarkblue}{rgb}{0,0.08,0.45}
\definecolor{wkblue}{rgb}{0.2, 0.3, 0.6}
\definecolor{meta-color}{rgb}{0.5, 0.5, 0.5}
\definecolor{bgblue}{RGB}{245,243,253}
\definecolor{ttblue}{RGB}{91,194,224}
\definecolor{mybrown}{RGB}{128,64,0}
\definecolor{titlecolor}{HTML}{4c9cff}
\definecolor{coolblue3}{rgb}{0.91, 0.94, 0.98}
\definecolor{myblue}{rgb}{0.9, 0.1, 0.94}
\definecolor{mygreen}{rgb}{0.64, 0.56, 0.88}
\definecolor{myyellow}{rgb}{0.68, 0.6, 0.1}
\definecolor{fancygreen}{rgb}{0.33, 0.68, 0.20}
\definecolor{salmon}{rgb}{0.94, 0.52, 0.49}
\definecolor{tablegreen}{rgb}{0.82, 0.94, 0.75}
\definecolor{tableblue}{rgb}{0.81, 0.90, 0.94}
\definecolor{tablered}{rgb}{0.97, 0.85, 0.85}
\definecolor{tableorange}{rgb}{0.96, 0.85, 0.81}

\usepackage{enumitem}
\usepackage{footnote}
\usepackage{lipsum}
\usepackage{setspace}

\usepackage{geometry}
\usepackage{fancyhdr}
\usepackage{lscape}
\usepackage{rotating}

\usepackage{algorithm}
\usepackage{algorithmic}

\usepackage{awesomebox}

\usepackage{natbib}
\usepackage{url}
\usepackage[colorlinks=true,linkcolor=mydarkblue,citecolor=mydarkblue,filecolor=mydarkblue,urlcolor=mydarkblue]{hyperref}

\usepackage{bbding}
\usepackage{imakeidx}
\makeindex
\usepackage[toc]{multitoc}
\usepackage[edges]{forest}
\usepackage[normalem]{ulem}
\usepackage{fontawesome5}
\usepackage{blindtext}
\usepackage{pgfplots}
\usepackage{pgfplotstable}
\pgfplotsset{compat=1.18}

\usepackage{listings}
\usepackage{listingsutf8}
\newcommand\JSONnumbervaluestyle{\color{blue}}
\newcommand\JSONstringvaluestyle{\color{red}}

\newif\ifcolonfoundonthisline

\makeatletter
\lstdefinestyle{json}
{
  showstringspaces    = false,
  keywords            = {false,true},
  alsoletter          = 0123456789.,
  morestring          = [s]{"}{"},
  stringstyle         = \ifcolonfoundonthisline\JSONstringvaluestyle\fi,
  MoreSelectCharTable =%
    \lst@DefSaveDef{`:}\colon@json{\processColon@json},
  basicstyle          = \ttfamily,
  keywordstyle        = \ttfamily\bfseries,
}

\newcommand\processColon@json{%
  \colon@json%
  \ifnum\lst@mode=\lst@Pmode%
    \global\colonfoundonthislinetrue%
  \fi
}

\lst@AddToHook{Output}{%
  \ifcolonfoundonthisline%
    \ifnum\lst@mode=\lst@Pmode%
      \def\lst@thestyle{\JSONnumbervaluestyle}%
    \fi
  \fi
  \lsthk@DetectKeywords%
}

\lst@AddToHook{EOL}%
  {\global\colonfoundonthislinefalse}
\makeatother

\newtcolorbox{myboxi}[1][]{
  breakable,
  title=#1,
  colback=red!5,
  colbacktitle=red!5,
  coltitle=black,
  fonttitle=\bfseries,
  bottomrule=0pt,
  toprule=0pt,
  leftrule=2pt,
  rightrule=2pt,
  titlerule=0pt,
  arc=0pt,
  outer arc=0pt,
  colframe=red,
}

\newtcolorbox{myboxnote}[1][]{
  breakable,
  title=#1,
  colback=orange!0,
  colbacktitle=orange!0,
  coltitle=black,
  fonttitle=\bfseries,
  bottomrule=0pt,
  toprule=0pt,
  leftrule=2pt,
  rightrule=2pt,
  titlerule=0pt,
  arc=0pt,
  outer arc=0pt,
  colframe=orange,
}

\definecolor{brightblue}{RGB}{33, 102, 172}
\usepackage{pifont}
\DeclareCaptionFont{black}{\color{black}}

\newenvironment{itemize*}%
 {\leftmargini=10pt\begin{itemize}%
  \setlength{\itemsep}{0pt}%
  \setlength{\parskip}{0pt}%
  }%
 {\end{itemize}}
\newenvironment{enumerate*}%
 {\begin{enumerate}%
  \setlength{\itemsep}{0pt}%
  \setlength{\parskip}{0pt}}%
 {\end{enumerate}}

\usepackage{etoolbox}
\newcounter{bibcount}
\makeatletter
\patchcmd{\@lbibitem}{\item[}{\item[\hfil\stepcounter{bibcount}{[\thebibcount]}}{}{}
\renewcommand\NAT@bibsetup%
  [1]{\setlength{\leftmargin}{\bibhang}\setlength{\itemindent}{-\parindent}%
      \setlength{\itemsep}{\bibsep}\setlength{\parsep}{\z@}}
\makeatother

\definecolor{myblue}{rgb}{0.9, 0.1, 0.94}
\definecolor{mygreen}{rgb}{0.64, 0.56, 0.88}
\definecolor{myyellow}{rgb}{0.68, 0.6, 0.1}
\definecolor{fancygreen}{rgb}{0.33, 0.68, 0.20}
\definecolor{salmon}{rgb}{0.94, 0.52, 0.49}
\definecolor{tablegreen}{rgb}{0.82, 0.94, 0.75}
\definecolor{tableblue}{rgb}{0.81, 0.90, 0.94}
\definecolor{tablered}{rgb}{0.97, 0.85, 0.85}
\definecolor{tableorange}{rgb}{0.96, 0.85, 0.81}

\newcommand{\model}{daVinci-kernel}

\begin{document}

%% ============================================================
%% Title and Authors
%% ============================================================

% Efficient Large-Scale Environment Synthesis: Methods, Data, and Scaling Laws
% Scaling Synthetic Environments: Methods, Data, and Scaling Laws
% Scaling Synthetic Environments: Methods, Data, and Scaling Laws
% OpenSWE: Efficient Environment Synthesis at Scale

% \title{OpenSWE: Open SWE Environment Synthesis at Scale}

% \title{OpenSWE: Efficient SWE Environment Synthesis at Scale}

\title{daVinci-kernel: Co-Evolving Skill Selection, Summarization, and Utilization via RL for GPU Kernel Optimization}

% Author list - retain lab affiliations
% \author[1,2,5]{Author One\textsuperscript{*}}
% \author[1,5]{Author Two\textsuperscript{*}}
% \author[3,5]{Author Three}
% \author[4,2,5]{Author Four}
% \author[1,2,5]{Pengfei Liu\textsuperscript{†}}
% \affil{\textsuperscript{1}SII  \quad \textsuperscript{2}SJTU  \quad \textsuperscript{3}GAIR}

\author[1,2,3]{Dayuan Fu}
\author[1,2]{Mohan Jiang}
\author[1]{Tongyu Wang}
\author[1]{Dian Yang}
\author[1]{Jiarui Hu}
\author[1]{Liming Liu}
\author[1]{Jinlong Hou}
\author[1,2,3]{Pengfei Liu
\textsuperscript{†}}
\affil{\textsuperscript{1}SII  \quad \textsuperscript{2}SJTU  \quad \textsuperscript{3}GAIR}

% \author{
% Dayuan Fu$^{1,2,3}$,
% Mohan Jiang$^{1,2}$,
% Tongyu Wang$^{1}$,
% Dian Yang$^{1}$,
% Jiarui Hu$^{1}$ \\
% Liming Liu$^{1}$,
% Jinlong Hou$^{1}$,
% Pengfei Liu$^{1,2,3}$ \textsuperscript{†}
% \\ \;
% $^{1}$SII \quad $^{2}$SJTU \quad $^{3}$GAIR
% }

% \author[2,3]{Shenyu Wu\textsuperscript{*}}
% \author[2,3]{Yunze Wu\textsuperscript{*}}
% \author[2,3]{Zerui Peng\textsuperscript{*}}
% \author[2,3]{Yaxing Huang\textsuperscript{*}}
% \author[1]{Jie Sun\textsuperscript{*}}
% \author[2,3]{Ji Zeng}
% \author[1,2]{Mohan Jiang}
% \author[2,3]{Lin Zhang}
% \author[2]{Yukun Li} 
% \author[1]{Jiarui Hu}
% \author[1]{Liming Liu}
% \author[1]{Jinlong Hou\textsuperscript{†}}
% \author[1,2,3]{Pengfei Liu\textsuperscript{†}}
% \affil{SII \quad \textsuperscript{2}SJTU \quad \textsuperscript{3}GAIR}
% \footnotetext[1]{* Equal contribution.}
\footnotetext[1]{† Corresponding authors.}

\maketitle

%% ============================================================
%% Header Configuration for Title Page
%% ============================================================
\pagestyle{fancy}
\thispagestyle{fancy}
\fancyhead{}
\lhead{
  \raisebox{-0.3cm}{\includegraphics[height=0.95cm]{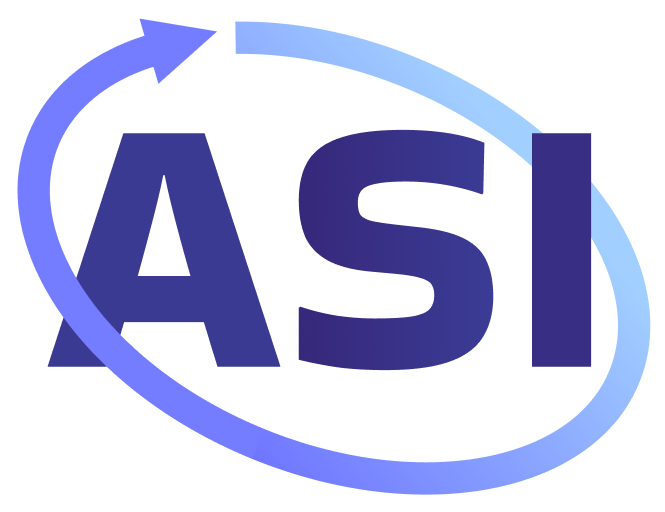}}
}
\rhead{%
  \raisebox{-0.2cm}{\includegraphics[height=0.7cm]{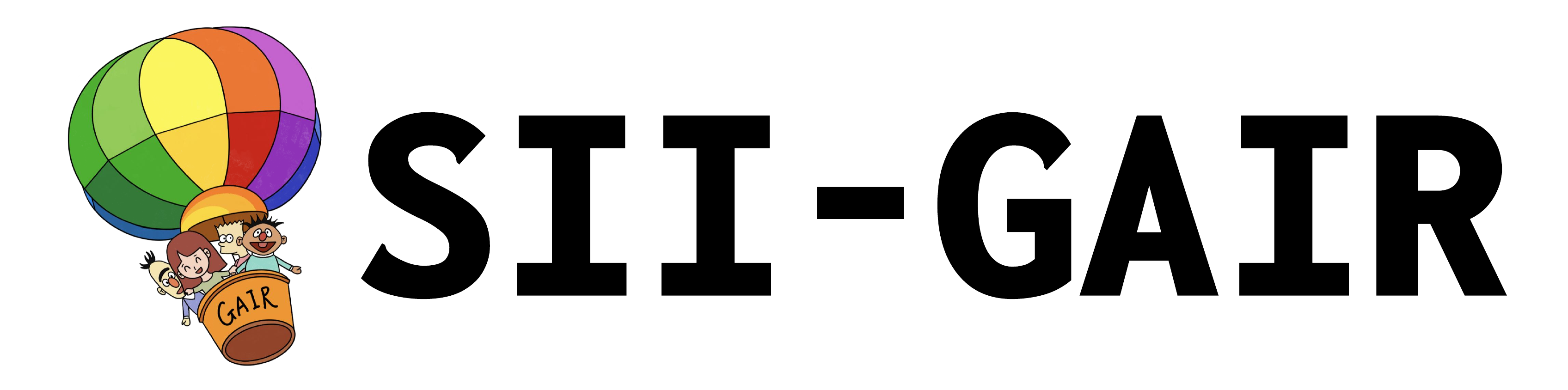}}%
}
\renewcommand{\headrulewidth}{0pt}
\setlength{\headsep}{2mm}

%% ============================================================
%% Author Footnotes
%% ============================================================
\begin{NoHyper}
\renewcommand{\thefootnote}{}
% \footnotetext{* Equal contribution.}
% \footnotetext{† Corresponding author.}
\end{NoHyper}
\vspace{-20pt}

%% ============================================================
%% Resource Links (customize as needed)
%% ============================================================
% {\centering
% \href{https://github.com/GAIR-NLP}{\raisebox{-.15em}{\includegraphics[height=1em]{assets/sii.png}}\ SII Open Source:}
% \quad \href{https://github.com/GAIR-NLP}{\textcolor{black}\faGithub\ Code}
% \quad \href{https://huggingface.co/GAIR}{\raisebox{-.15em}{\includegraphics[height=1em]{figures/huggingface-icon.png}}\ Models}
% \quad \href{https://huggingface.co/datasets/GAIR}{\textcolor{violet}\faDatabase\ Datasets}
% \par}

\vspace{20pt}

%% ============================================================
%% Main Content
%% ============================================================
% !TEX root = main.tex
% maintext.tex - Main content of the paper
% This file contains all sections of the paper

%% ============================================================
%% Abstract
%% ============================================================
\begin{abstract}
GPU kernel optimization represents a paradigm where functional correctness is assumed and execution efficiency is the objective---demanding not only capable code generation, but the ability to maintain and evolve optimization knowledge as the model's capability frontier continuously advances. We present \textbf{\model{}}, a reinforcement learning framework that couples skill discovery with skill exploitation through a dynamically evolving skill library. \model{} jointly trains three agents sharing one LLM backbone: a Skill Selection Agent that retrieves relevant techniques via BM25 and LLM reranking, a Policy Agent that generates multi-turn CUDA/Triton kernels conditioned on selected skills, and a Skill Summary Agent that distills successful rollouts into reusable skills. Candidate skills are added only after execution-based verification confirms reproducible speedups. All three agents share a single LLM backbone, are initialized via a structured SFT cold start on diversity-filtered data, and are then jointly optimized end-to-end with multi-turn REINFORCE and per-agent advantage estimation. On KernelBench, \model{}-14B achieves 37.2\%, 70.6\%, and 32.2\% on Level~1, Level~2, and Level~3 under the Fast$_1$ threshold, outperforming the strongest prior RL-trained model, Dr.~Kernel-14B. 
The gains are especially pronounced on harder tasks, suggesting that evolving and reusing optimization skills becomes increasingly important as kernel optimization requires deeper and more coordinated transformations. 
Comprehensive ablation studies further show that removing inference-time skill injection causes a large performance drop, especially on difficult tasks, while training without joint selection and summarization weakens high-threshold speedups. 
Replacing LLM-based selection with BM25-only retrieval can still help find shallow speedups, but fails to reliably support more demanding optimization thresholds. 
These results demonstrate that skill injection, skill selection, skill summarization, and diversity-aware initialization provide complementary benefits, validating the need for co-evolving skill-policy optimization in GPU kernel generation. The code can be found in \url{https://github.com/GAIR-NLP/daVinci-kernel}
\end{abstract}

\begin{figure*}[htbp]
    \centering 
    \includegraphics[width=0.98\textwidth]{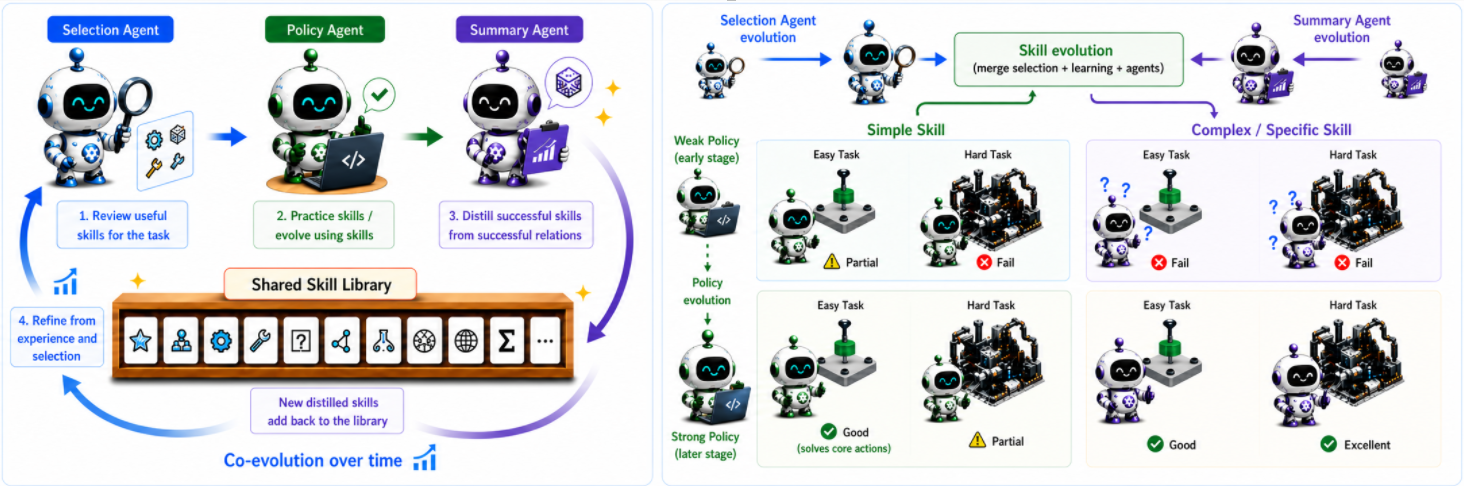} 
    \caption{\textbf{Skill-policy co-evolution in \model{}.}
The selection, policy, and summary agents form a closed RL loop, where task-relevant skills guide optimization and successful rollouts are distilled back into reusable skills. As the policy improves, the useful skill frontier shifts from simple skills to more complex and task-specific skills, enabling increasingly difficult kernel optimization.}
    \label{fig:teaser}
    \vspace{-3mm}
\end{figure*}

% \begin{figure}[h]
%     \centering
%     \includegraphics[width=0.93\linewidth]{figures/teaser8.pdf}
%     \caption{Image construction and performance overview of OpenSWE.}
%     \label{fig:teaser}
% \end{figure}

\newpage
\pagestyle{fancy}
\lhead{\rightmark}
\renewcommand{\headrulewidth}{0.7pt}
\setlength{\headsep}{5mm}
\clearpage

\newpage
% Reset footnote numbering for main content
\renewcommand{\thefootnote}{\arabic{footnote}}
\setcounter{footnote}{0}

%% ============================================================
%% Introduction
%% ============================================================

\section{Introduction}
\label{sec:introduction}

%如果后面又要加新东西太长的话就删删

Large language models are increasingly shifting code generation from one-shot program synthesis toward feedback-driven program optimization~\citep{jimenez2023swe,fu2026davinci}. GPU kernel optimization~\citep{ouyang2025kernelbench,li2025tritonbench} is a particularly representative setting in this transition. Unlike many code-generation tasks whose primary objective is functional correctness, kernel optimization typically starts from an implementation that is already correct but inefficient, and improves it through a sequence of transformations. The resulting improvements are measurable, incremental, and compositional, rather than binary~\citep{liu2026dr}. 
% Consequently, the learning signal in kernel optimization is not naturally expressed as success or failure, but as a continuous reward landscape defined by end-to-end execution efficiency.

This progressive structure makes reinforcement learning a natural paradigm, but it also introduces a more fundamental difficulty~\citep{liu2026dr,li2025autotriton,guo2026drtriton,woo2025tritonrl}. As the policy improves, the source of further gains changes. Early in training, substantial improvements often come from broad transformations that correct obvious inefficiencies. Later, however, the remaining performance gap is increasingly concentrated in subtler, and more tightly coupled optimization decisions. Therefore, the knowledge gained from the early stage may useless in the next stage.
% Kernel optimization is therefore governed by a moving capability frontier: the knowledge that is most useful at one stage of training may no longer be the primary bottleneck at a later stage, while new forms of knowledge become necessary for further improvement. The challenge is thus not only to optimize under continuous rewards, but to sustain learning when the value of optimization knowledge is itself non-stationary.
This observation suggests that continual improvement cannot rely on stable knowledge library. A learning system must also be able to preserve optimization knowledge discovered at earlier stages, re-evaluate its usefulness as the policy evolves, and expose the most relevant knowledge to future decisions. 
% In kernel optimization, this requirement is especially pronounced because the utility of prior experience is determined not by abstract similarity alone, but by whether it still helps close the current performance gap under real execution. 
What is needed, therefore, is not merely a stronger optimizer for the current task, but a mechanism for maintaining and reusing optimization knowledge in a way that remains aligned with the agent's evolving frontier.

Existing approaches address important parts of this problem. On the one hand, recent kernel agents have substantially improved optimization quality through prompting, execution-guided refinement, post-training, and reinforcement learning, thereby strengthening the generator and the search process for the current task~\citep{liu2026dr,li2025autotriton,guo2026drtriton,woo2025tritonrl,cao2026k,wei2025astra,zhu2026qimeng}. On the other hand, a growing body of work on skill-based agents and reinforcement learning has shown that experience can be abstracted into reusable units that support transfer across tasks and training stages~\citep{lu2026skill0,xia2026skillrl,li2026arise,tu2026dynamic}. However, in kernel optimization, such reusable units must satisfy two additional requirements. First, their usefulness must be judged relative to the policy's current competence, since techniques that were once beneficial may later become redundant after being internalized. Second, newly acquired knowledge should not be retained on plausibility alone, but on whether it yields verifiable execution gains in the target environment. These requirements shift the problem from generic experience accumulation to \emph{frontier-aware skill evolution}: reusable skills must be continually selected, and summarized throughout training.

Motivated by this perspective, we propose \model{}, a reinforcement learning framework for GPU kernel optimization that closes the loop between skill discovery and skill exploitation. \model{} jointly trains a selection mechanism that retrieves task-relevant skills, a policy that performs multi-turn kernel optimization under skills, and a summary agent that distills successful rollouts into candidate skills.
% Crucially, candidate skills are retained only after execution-grounded verification demonstrates reproducible gains in the target environment.
In this way, \model{} treats skill maintenance not as an auxiliary component attached to optimization, but as part of the learning problem itself. Empirically, we show that \model{}-14B achieves 37.2\%, 70.6\%, and 32.2\% on Level~1, Level~2, and Level~3 under the Fast$_1$ threshold outperforming Dr.~Kernel-14B~\citep{liu2026dr} by up to 46\% proves the effectness of our method. The following ablation studies further validate the contribution of each component.
% and surpassing frontier closed-source models including GPT-5 and Claude-4.5-Sonnet with a 14B open model---and that its dynamic skill evolution mechanism substantially improves both knowledge reuse and downstream kernel optimization quality.

In summary, our contributions are as follows:
% \begin{itemize}[leftmargin=10pt,itemsep=1pt,topsep=1pt]
    % \item We identify the non-stationary utility of optimization knowledge as a central challenge in reinforcement learning for GPU kernel optimization, and argue that the main difficulty lies not only in optimizing under continuous rewards, but also in maintaining alignment between reusable knowledge and the policy's evolving capability frontier.
\begin{itemize}[leftmargin=10pt,itemsep=1pt,topsep=1pt]
    \item We propose \model{}, a unified framework that couples skill selection, skill utilization, and skill summarization into a single reinforcement learning loop to improve agent's performance.
    \item Empirically, we show that \model{}-14B achieves 37.2\%, 70.6\%, and 32.2\% on Level~1, Level~2, and Level~3 under the Fast$_1$ threshold on KernelBench, outperforming Dr.~Kernel-14B by up to 46\% on Level~3. These results demonstrate the effectiveness of \model{}.
    \item We conduct comprehensive ablation studies to isolate the contribution of each component. The results show that inference-time skill injection is critical for difficult tasks, joint multi-agent RL prevents mismatch between the policy and the evolving skill library, LLM-based skill selection improves high-quality speedups, and diversity-filtered SFT data mitigates skill mode collapse.
\end{itemize}

\section{Related Work}
\label{sec:related}

\textbf{Kernel Agents.} 
Another closely related line of work studies the use of LLMs for GPU kernel generation and optimization. Early approaches mainly rely on prompting, iterative refinement, or execution-guided search. For example, CUDA-LLM~\citep{chen2025cuda} improves kernels using compilation and runtime feedback, while GPU Kernel Scientist formulates optimization as repeated proposal, execution, and revision. 
More recent work has moved toward dedicated post-training and reinforcement learning recipes, including Dr.\ Kernel~\citep{liu2026dr}, DRTriton~\citep{guo2026drtriton}, and CUDA Agent~\citep{dai2026cuda}, which improve generation quality through supervised post-training, reinforcement learning, multi-turn optimization, large-scale synthetic data, and more robust training objectives. In parallel, works such as Kernel-Smith~\citep{du2026kernel} show that execution feedback, evolutionary search, and post-training optimization can substantially improve the practical performance of generated kernels. 
While these approaches have significantly advanced LLM-based kernel optimization, they focus mainly on improving the generator or the search process for the current task, rather than on maintaining a skill library whose contents are continually regenerated, reselected, and execution-validated as the agent evolves.

\textbf{Skill-based reinforcement learning.} 
A growing body of work studies how agents can accumulate reusable skills from past experience, rather than solving each task from scratch. Early systems such as Voyager~\citep{wang2023voyager} and ExpeL~\citep{zhao2024expel} demonstrate that explicitly storing experience or insights in an external repository can substantially improve long-horizon planning and cross-task transfer. Building on this idea, recent work has integrated skill learning more tightly with reinforcement learning. SAGE~\citep{liang2025sage} combines skill libraries with RL, places greater emphasis on sequential rollouts, and skill-integrated rewards. SkillRL~\citep{xia2026skillrl} constructs a hierarchical skill bank through experience distillation and allows the skill library to co-evolve recursively with the policy during RL. In contrast, SKILL0~\citep{lu2026skill0} focuses less on maintaining an external skill library throughout training, and instead studies how retrieved skills can be progressively internalized into model parameters through a curriculum that gradually removes skill context. However, they largely treat skill utility as relatively static. In practice, skill value changes during training: some skills become absorbed into the model and no longer need to be injected, while stronger agents continuously encounter new capability frontiers that require new knowledge. Thus, for problems such as GPU kernel optimization, the key is not simply to accumulate skills, but to let skill generation, selection, and updating co-evolve with training, while retaining only reliably useful skills under execution-based verification.

\section{\model{}}

\begin{figure*}[htbp]
    \centering 
    \includegraphics[width=1.\textwidth]{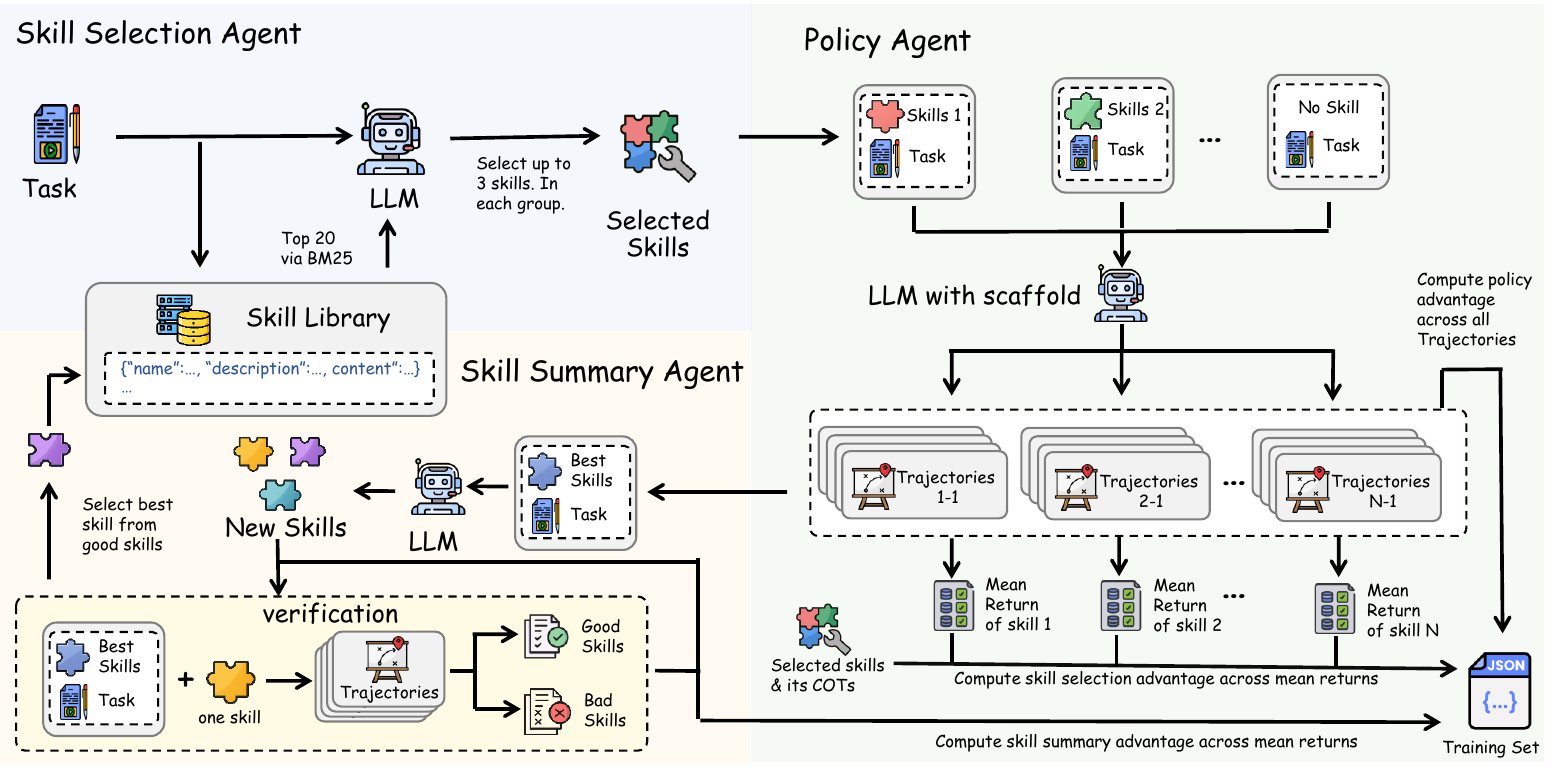} 
    \caption{The structure of \model{}.} 
    \label{fig:main}
\end{figure*}

We present \textbf{\model{}}, a reinforcement learning framework that trains an LLM to write optimized CUDA kernels while continuously accumulating and reusing a library of optimization techniques. The core idea is to close the loop between \emph{discovery} and \emph{exploitation}: when the model acting as a policy, it needs to discover effective optimization tricks and distilled into a reusable skill on subsequent tasks. The relevant skills are retrieved and injected into the model’s context, guiding the policy generation and enabling knowledge transfer across tasks and training steps. 
% Importantly, this process is implemented within a unified model and jointly optimized, so the framework not only maintains an external skill library but also encourages the model itself to internalize the ability to apply existing skills, generate improved new skills from experience, and summarize them into reusable abstractions.
The framework comprises three agents trained jointly with reinforcement learning. As illustrated in Figure \ref{fig:main}, these are: (i) a \textbf{Skill Selection Agent} that retrieves task-relevant skills from a growing library; (ii) a \textbf{Policy Agent} that generates multi-turn CUDA kernels, optionally conditioned on injected skills; and (iii) a \textbf{Skill Summary Agent} that distills successful rollouts into new skills. All three agents share the same LLM backbone and are trained end-to-end with the same relative objective.

\subsection{Preliminary}

\paragraph{Introduction about Kernel Optimization.}
Optimizing CUDA kernels is inherently a \emph{continuous} and iterative process, where a correct implementation can be progressively improved through a sequence of transformations (e.g., parallelization and fusion), each yielding incremental performance gains.
As a result, the objective is not binary, but \emph{graded} and accumulative across multiple steps, with partial improvements providing meaningful training signals.

In practice, this setting introduces several challenges.
The reward signal is often sparse and delayed, as substantial speedups typically require coordinated modifications over multiple turns.
At the same time, the optimization space contains many degenerate solutions, where the model produces syntactically valid but ineffective changes (``lazy optimization''), such as preserving the original implementation or applying transformations that do not impact end-to-end runtime.
Moreover, performance gains are tightly coupled to system-level bottlenecks, making it insufficient to optimize local code patterns without considering their global runtime contribution.

\paragraph{TRLOO}
\label{trloo}
Our policy agent training setup follows \textsc{Dr.\ Kernel}, adopting a unified reinforcement learning objective that combines \textbf{turn-level REINFORCE Leave-One-Out (TRLOO)} with mismatch control and profiling-aware optimization.
For each trajectory $i$ at turn $t$, a bottleneck-aware reward is defined as
\begin{equation}
R_{i,t}
=
C(y_{i,t}) \cdot \big(1 + \mathrm{speedup}_{i,t} + \mathrm{PR}_{i,t}\big),
\quad
\mathrm{PR}_{i,t}
=
\frac{T_{\text{generated}}}{T_{\text{total}}},
\end{equation}
where $C(\cdot)$ denotes correctness, and $\mathrm{PR}$ measures the fraction of end-to-end runtime covered by generated kernels.

The return is computed as $G_{i,t} = \sum_{t' \ge t} \gamma^{t'-t} R_{i,t'}$, and advantages are estimated using TRLOO:
\begin{equation}
A_{i,t}
=
G_{i,t}
-
\frac{1}{N_t - 1}
\sum_{j \neq i} G_{j,t}.
\end{equation}

To stabilize optimization, \textbf{MRS} filters samples with large importance ratio deviation:
\begin{equation}
w_i
=
\exp\!\left(
\frac{1}{|\mathcal{T}_i|}
\sum_{k}
\log
\frac{\pi_{\text{train}}(a_k \mid s_k)}
     {\pi_{\text{rollout}}(a_k \mid s_k)}
\right).
\end{equation}

Finally, \textbf{PRS} prioritizes optimization on kernels that impact end-to-end performance:
\begin{equation}
p_{i,t}
=
\mathrm{clip}\!\left(
\frac{\mathrm{PR}_{i,t} - \tau}{s},\, 0,\, 1
\right),
\end{equation}
and trajectories are sampled proportionally to $p_{i,t}$.

\subsection{Skill Library}

The skill library $\mathcal{L}$ is a collection of GPU optimization techniques.
Each skill records with five fields:
\begin{itemize}[leftmargin=10pt,itemsep=1pt,topsep=1pt]
  \item \textbf{\texttt{name}}: a short \texttt{snake\_case} identifier, used as a unique key within a snapshot and as the retrieval target returned by the Selection Agent's tool call.
  \item \textbf{\texttt{description}}: a one-sentence summary of what the technique does and when it applies, shown to the Selection Agent to support relevance judgement without loading full content.
  \item \textbf{\texttt{scope}}: a coarse category
        that helps the Selection Agent filter by kernel subtype.
  \item \textbf{\texttt{tags}}: a list of keywords
        used to describe the domain of the skill. 
  \item \textbf{\texttt{content}}: a free-text Markdown document with the full technique description, code snippets, and usage notes, which is the text ultimately injected into the policy's context.
\end{itemize}
The library is persisted as JSONL snapshots keyed by training step, enabling exact checkpoint restart.

% ---------------------------------------------------------------

\subsection{Skill Selection Agent}
\label{sec:selection}

Before the policy attempts a task, the Selection Agent acts as a \emph{retriever}: given the task description, it queries the skill library and assembles a concise set of potentially useful optimization techniques to provide as context. The key challenge is to surface the most relevant skills efficiently as $\mathcal{L}$ grows over training.
To this end, we use a two-stage procedure that combines lightweight lexical search with LLM-based relevance judgement.

% To support efficient retrieval, we maintain a BM25 index over the concatenation of
% each skill's \texttt{name}, \texttt{description}, \texttt{tags}, and \texttt{content}
% fields; the index is built lazily after each library load and rebuilt after each flush.

\textbf{Stage 1: BM25 pre-filtering.} 
To support efficient retrieval, we maintain a BM25 index~\citep{robertson2009probabilistic} over the concatenation of each skill's \texttt{name}, \texttt{description}, \texttt{tags}, and \texttt{content} fields. The index is built lazily after each library load and rebuilt after each training step's flush.
Given the task description, we retrieve the top-$N_{\mathrm{BM25}}$ candidate skills from $\mathcal{L}$ by BM25 score.
% This bounds the candidate set to a fixed size regardless of library cardinality, keeping the selection prompt length stable as $\mathcal{L}$ grows.

\textbf{Stage 2: LLM selection.} 
After the Selection Agent receives a list of the BM25-retrieved candidates together with the task description, we then use LLM to select at most $k_{\mathrm{select}}$ skill. Specifically, we will add the task description, skills \texttt{name}, and skills \texttt{description} to the system prompt and ask the model to call a tool named \texttt{select\_skills}, which takes a \texttt{think} field for reasoning and returns an ordered list of at most $k_{\mathrm{select}}$ skills, each with a \texttt{name} and a brief \texttt{reason}.
If the library is empty or no skill scores above the BM25 threshold, the agent returns an empty selection and $c = \varnothing$.

% ---------------------------------------------------------------
\subsection{Policy Agent}
\label{sec:policy}

The Policy Agent is the \emph{executor}: it receives the task and an optional skill context $c$ produced by the Selection Agent, and generates an optimized CUDA kernel
implementation through multi-turn dialogue with a code execution environment. Its core capability is to leverage the injected skills to write kernels.
When the Skill Selection Agent provides skills, the \texttt{content} and \texttt{name} fields of the skill will be injected into the task prompt before the agent begins generation. When $c = \varnothing$, the prompt is passed through unmodified.

\subsection{Skill Summary Agent}
\label{sec:summary}

The Summary Agent closes the loop by acting as a \emph{distiller}: when the Policy Agent produces a strong rollout, the Summary Agent analyzes the successful trajectory and extracts the underlying optimization insight as a new skill to be added to $\mathcal{L}$.
Over training, this mechanism turns individual task successes into reusable knowledge, continuously enriching the library from which future selection and policy calls benefit.
The agent operates conditionally and with quality controls to ensure that only genuinely transferable techniques enter the library.

\textbf{Trigger condition.} 
The Summary Agent is invoked on a completed policy rollout only when the rollout exhibits meaningful improvement:
\begin{equation}
  R^{*} > \alpha \cdot r_1 \quad \text{and} \quad R^{*} > \beta,
  \label{eq:trigger}
\end{equation}
where $R^{*}$ is the maximum speedup across all turns within a single rollout, $r_1$ is the first-turn speedup of that same rollout, and $\alpha$, $\beta$ are relative and absolute thresholds, with $\alpha$ ensuring the improvement is sufficiently significant rather than arising from evaluation noise or fluctuation, and $\beta$ imposing an absolute floor to prevent lazy optimization, i.e., trivial or negligible gains.
By comparing the improvement between the best turn and the first turn, the agent can find the real improvement signal beyond the task or the inherent skills.

\textbf{Summary inference.} 
For a triggered rollout, the Summary Agent receives (1) The system prompt about summarization. (2) The original task context (3) The skill content injected into the prompt (or a note that no skill was provided). (4) The first turn kernel code
(5) The best turn kernel code (6) The \texttt{update\_skill\_library} tool content, and it calls the \texttt{update\_skill\_library} tool exactly once, proposing at most $S_{\max}$ new skills.

A lazy skill filter then use a rule-based method to reject any proposed skill whose \texttt{name},
\texttt{description}, \texttt{tag} or \texttt{content} suggests bypassing custom CUDA/Triton
kernels (e.g., \textit{``fall back to PyTorch''}, \textit{``use \texttt{torch.compile}''}),
ensuring only genuine kernel optimization techniques enter the library.

\textbf{Skill verification.} 
Each candidate skill is validated by re-running the policy on the original task with the skill injected into the system prompt. 
A skill is added to the library only if the resulting first-turn speedup satisfies:
\begin{equation}
  r_{\mathrm{verify}} \;\geq\; \max\!\bigl(\beta,\; \alpha \cdot r_1\bigr),
  \label{eq:verify}
\end{equation}
where $r_{\mathrm{verify}}$ denotes the first-turn speedup achieved with both the original skills and the candidate skill applied, and $r_1$ is the first-turn speedup of the triggering rollout that with only the original skills.

\subsection{SFT cold start}
\label{sec:sft}

Before reinforcement learning, we initialize all three agents through supervised fine-tuning.
The training data for each agent is constructed offline from a shared seed set of multi-turn kernel optimization trajectories collected from \texttt{CUDALLM} tasks~\citep{chen2025cuda} using a strong LLM like GPT-5.4 as the policy.

\textbf{Summary SFT.} 
Summary Agent training data is constructed in three phases.
In Phase~1, we apply Eq.~\ref{eq:trigger} to the Dr.\ Kernel cold-start dataset to filter trajectories that exhibit sufficiently strong improvement signals, ensuring that only high-quality candidates are retained for skill extraction.
In Phase~2, we randomly select half of the filtered trajectories and run the Summary Agent on each to generate \texttt{update\_skill\_library} tool-call demonstrations. The resulting skills are aggregated to form an initial skill library, which serves as the foundation for subsequent SFT stages.
In Phase~3, we process the remaining half using BM25 to retrieve skills from the Phase~2 library and inject them into the prompt, producing skill-conditioned summarization demonstrations.
The tool calls in Phase~2 and Phase~3 will be transformed into the supervised training example.

\textbf{Selection SFT.} 
After the seed skills are generated by last phases, we use the same dataset to generate the Selection SFT data. For each seed task, we run the Selection Agent call using the two-stage BM25+LLM procedure.
Calls are executed against the skill library in the Summary SFT generation stage.
Each call produces one \texttt{select\_skills} tool-call conversation as the SFT data.

\textbf{Policy SFT.} 
We use two data sources for the policy.
The \emph{base policy SFT} data comes directly from the Dr.\ Kernel coldstart dataset.
The \emph{skill-injected policy SFT} data is derived from the same base tasks but with skill content injected into the first user turn.
The skill is selected from the whole skill library generated from the summary agent above via BM25.
We use GPT-5.4 to sample the trajectories with real kernel environment feedbacks for 5 turns each task, align with the setting in \emph{base policy SFT}. The two sources are combined as the final Policy SFT data, so the model retains its baseline kernel-writing capability while also learning the skill-conditioned format.

\textbf{Filtering.} 
After constructing the skill summarization and skill selection datasets, we observe a pronounced distributional imbalance: the model often generates overly generic and easily learnable skills, such as \textit{identifying and optimizing hot paths}. These recurring patterns can accelerate early-stage RL results but later may reduce skill diversity and limit further improvement. To mitigate this, we apply structured downsampling to both datasets. We encode all skill texts with \texttt{Qwen3-Embedding-8B} and cluster the resulting representations using HDBSCAN. For the skill summarization dataset, we iteratively downsample each cluster until at most two instances remain, yielding a more uniform skill distribution and reducing the dominance of generic patterns. For the skill selection dataset, we perform cluster-aware deduplication within each tool call: skills from the same semantic cluster are pruned to the most task-relevant representative, prioritizing alignment with the \texttt{think} reasoning field. Trajectories whose reasoning becomes inconsistent after pruning are discarded.

We shuffled all 3 types of datas after downsampling to train the cold start model.

\subsection{Joint Reinforcement Learning}
\label{sec:rl}

\textbf{Training setup.} 
For each task in a training batch of size $B$, we expand to $k{+}1$
parallel \emph{schemes}.
The first $k$ schemes are obtained by running $k$ independent Selection Agent calls
on $\mathbf{m}$, each following the two-stage BM25+LLM procedure
described in \S\ref{sec:selection} and producing a distinct skill context
$c_1,\ldots,c_k$.
The $(k{+}1)$-th scheme is the \emph{null scheme} ($c = \varnothing$), which always
serves as an unmodified baseline.
For each scheme $i$, the Policy Agent generates $n$ rollout samples, yielding
trajectories $\tau^{\mathrm{pol}}_{i,j}$ with reward $R_{i,j,t}$ and returns $G_{i,j,t}$ at each turn in the trajectories.
For triggered tasks, we select the best policy trajectories from each task as the summary seed and run $s$ independent Summary Agent calls.
All agents are trained jointly using REINFORCE with the advantage below:

\textbf{Policy advantage.} 
We use multi-turn TRLOO, grouping all $(k{+}1){\times}n$ rollouts for
the same task into a single LOO group.
This treats different skill schemes as parallel rollout samples, yielding a
lower-variance baseline than per-scheme grouping:
\begin{equation}
  A^{\mathrm{pol}}_{i,j,t}
  \;=\;
  G_{i,j,t}
  \;-\;
  \frac{1}{(k{+}1)n - 1}
  \sum_{(i',j') \neq (i,j)} G_{i',j',t}.
  \label{eq:pol_adv}
\end{equation}

\textbf{Selection advantage.} 
The reward for scheme $i$ is set to the mean policy return under that scheme,
$\bar{R}_i = \frac{1}{n}\sum_j \max\limits_{t} R_{i,j,t}$.
LOO is then applied across the $k$ non-null schemes within the same task:
\begin{equation}
  A^{\mathrm{sel}}_i
  \;=\;
  \bar{R}_i
  \;-\;
  \frac{1}{k}
  \sum_{i' \neq i} \bar{R}_{i'}.
  \label{eq:sel_adv}
\end{equation}
This trains the Selection Agent to prefer skill combinations that consistently improve
policy returns over alternative selections on the same task.

\textbf{Summary advantage.} 
For each summary call $m$, we define its reward as the verification speedup of the best generated skill,
\[
  R^{\mathrm{sum}}_m
\;=\;
r^{(m)}_{\mathrm{verify}}
\cdot
\mathbf{1}\!\left[
r^{(m)}_{\mathrm{verify}} \;\ge\; \max\!\bigl(\beta,\; \alpha \cdot r^{(m)}_1\bigr)
\right]
\]
If all proposed skills corresponding to the task-original skill pair fail verification, all the summary result under that pair will be removed.
We then apply LOO within the $s$ parallel summary calls for the same task:
\begin{equation}
  A^{\mathrm{sum}}_m
  \;=\;
  R^{\mathrm{sum}}_m
  \;-\;
  \frac{1}{s - 1}
  \sum_{m' \neq m} R^{\mathrm{sum}}_{m'}.
  \label{eq:sum_adv}
\end{equation}
This encourages the Summary Agent to generate skills that achieve higher verified speedup compared to alternative summaries on the same task.

\textbf{Combined objective.} 
The final training loss combines all three agent types with per-agent weights:
\begin{equation}
  {Loss}
  \;=\;
  {Loss}^{\mathrm{policy}}
  \;+\; w_{\mathrm{select}}\,{Loss}^{\mathrm{select}}
  \;+\; w_{\mathrm{summary}}\,{Loss}^{\mathrm{summary}}.
  \label{eq:loss}
\end{equation}

\textbf{Skill library update.} 
After all $s$ summary calls for a triggered task complete, we select the verified skill with the highest $r_{\mathrm{verify}}$ across all $s$ skill to enter the library cache $\mathcal{L}_{cache}$. If no skill passes verification, nothing is staged.
At the end of each training step, all staged skills in $\mathcal{L}_{cache}$ are atomically flushed to a new versioned snapshot of $\mathcal{L}$ and broadcast to all rollout workers, so every subsequent step immediately benefits from the newly acquired knowledge. The $\mathcal{L}_{cache}$ is then cleared.
This one-per-task rule keeps library growth conservative and ensures that only the most reliably useful technique from each trajectory is retained.

\section{Experiment}

\subsection{Experimental Setup}

\textbf{Benchmark and metrics.} 
We evaluate \model{} on the KernelBench~\citep{ouyang2025kernelbench}, which organizes GPU kernel optimization tasks into three difficulty levels.
Each task provides a reference PyTorch implementation; the model must produce a functionally correct Triton kernel that outperforms the baseline in end-to-end wall-clock time.
We report performance as the percentage of tasks achieving at least $x\times$ speedup over the reference PyTorch implementation, denoted \mbox{Fast$_x$}, for thresholds $x \in \{1.0, 1.2, 1.5, 2.0\}$.
 Stricter thresholds capture qualitatively different optimization quality: Fast$_1$ measures whether the model produces \emph{any} custom kernel speedup, while Fast$_1.2$ requires that the model consistently identify and exploit the dominant performance bottleneck.

\textbf{Baselines.} 
We compare against three categories:
(i)~\emph{Frontier closed-source models} used in zero-shot settings: GPT-5, Claude-4.5-Sonnet, Deepseek-V3.2-Thinking, and GLM-4.7~\citep{liu2025deepseek,zeng2025glm};
(ii)~\emph{Open-weight models without task-specific training}: Qwen3-8B, Qwen3-32B, Qwen3-Coder-A30BA3, and AutoTriton~\citep{li2025autotriton,yang2025qwen3};
(iii)~\emph{Domain-specifically trained models}: Dr.~Kernel, the strongest published RL-trained kernel optimization baselines under 8B/14B settings.

\begin{table}[t]
\centering
\resizebox{\textwidth}{!}{
\begin{tabular}{lcccccccccccc}
\toprule
\textbf{Model} 
& \multicolumn{4}{c}{\textbf{LEVEL1}} 
& \multicolumn{4}{c}{\textbf{LEVEL2}} 
& \multicolumn{4}{c}{\textbf{LEVEL3}} \\
\cmidrule(lr){2-5} \cmidrule(lr){6-9} \cmidrule(lr){10-13}
& Fast$_1$ & Fast$_{1.2}$ & Fast$_{1.5}$ & Fast$_2$
& Fast$_1$ & Fast$_{1.2}$ & Fast$_{1.5}$ & Fast$_2$
& Fast$_1$ & Fast$_{1.2}$ & Fast$_{1.5}$ & Fast$_2$ \\
\midrule

GPT-5*                & 19.5 & 16.5 & 12.5 & 11.0 & 46.7 & 28.6 & 13.1 & 3.0 & 21.0 & 12.0 & 4.0 & 2.0 \\
Claude-4.5-Sonnet*    & 15.5 & 13.5 & 11.0 & 8.5  & 50.0 & 26.7 & 9.2  & 1.8 & 21.0 & 11.0 & 5.0 & 4.0 \\
Deepseek-V3.2-Thinking* & 7.5 & 5.5 & 4.5 & 4.0  & 11.0 & 6.5  & 2.5  & 0.5 & 2.0  & 1.0  & 0.0 & 0.0 \\
GLM-4.7*              & 19.4 & 17.2 & 13.1 & 10.4 & 30.0 & 20.5 & 8.5  & 3.5 & 5.0  & 2.0  & 2.0 & 2.0 \\
Qwen3-8B*             & 5.8  & 4.8  & 4.1  & 3.4  & 13.0 & 5.6  & 2.0  & 1.1 & 5.7  & 0.2  & 0.0 & 0.0 \\
Qwen3-32B*            & 6.1  & 4.9  & 4.3  & 4.0  & 14.0 & 9.4  & 2.4  & 0.2 & 3.5  & 0.0  & 0.0 & 0.0 \\
Qwen3-Coder-A30BA3*   & 6.0  & 5.2  & 5.1  & 3.8  & 12.6 & 5.0  & 1.5  & 0.3 & 7.0  & 1.0  & 0.0 & 0.0 \\
AutoTriton*           & 4.5  & 3.6  & 2.8  & 2.1  & 30.6 & 9.2  & 2.6  & 0.5 & 7.5  & 0.0  & 0.0 & 0.0 \\
Cold-Start-8B*       & 7.5  & 6.6  & 5.0  & 4.3  & 8.8  & 5.6  & 1.8  & 0.4 & 0.5  & 0.0  & 0.0 & 0.0 \\
Cold-Start-8B (reproduce)       & 11.4 & 9.3 & 7.4 & 4.9  & 11.8 & 8.6 & 5.9 & 3.5  & 0.5 & 0.3 & 0.3 & 0.0 \\
Cold-Start-14B (reproduce)      & 13.6 & 11.6 & 8.5 & 6.4  & 20.9 & 12.4 & 7.9 & 4.1 & 2.0 & 0.8 & 0.3 & 0.3 \\
Dr. Kernel-8B*  & 15.9 & 12.8 & 10.9 & 8.4  & 46.0 & 20.0 & 5.0  & 1.5 & 10.8 & 1.0  & 0.0 & 0.0 \\
Dr. Kernel-8B  (reproduce)   & 17.9 & 14.9 & 10.5 & 7.2  & 26.8 & 9.4 & 5.2 & 2.0 & 3.0 & 0.5 & 0.5 & 0.5 \\
Dr. Kernel-14B*     & 20.3 & 16.9 & 13.2 & 11.6  & 49.2 & 25.6 & 7.4  & 2.1 & 8.8  & 1.2  & 0.2 & 0.0 \\
Dr. Kernel-14B  (reproduce)  & 30.4 & 26.0 & 20.4 & 13.9 & 58.5 & 18.1 & 7.5 & 4.1 & 22.1 & 2.8 & 1.8 & \textbf{1.5 }  \\
\midrule
\model{}-8B   & 26.1 & 20.4 & 16.8 & 13.0 & 44.8 & 22.1 & \textbf{13.1} & \textbf{7.9} & 10.1 & 3.0 & 2.0 & 0.8  \\
\model{}-14B   & \textbf{37.2} & \textbf{27.3} & \textbf{22.9} & \textbf{16.2} & \textbf{70.6} & \textbf{27.1} & 11.4 & 6.2 & \textbf{32.2} & \textbf{7.3} & \textbf{2.3} & 1.3 \\
\bottomrule
\end{tabular}
}
\caption{Main benchmark results across three difficulty levels and four speedup thresholds. Results marked $^*$ are from the Dr.~Kernel paper. Bold denotes the best result in each column.}
\label{tab:main}
\end{table}

\textbf{Training setup.} 
\model{} is built on Qwen3-8B-Base and Qwen3-14B-Base backbones~\citep{yang2025qwen3}. All three agents share weights and are jointly fine-tuned.
The SFT cold start uses the Dr.\ Kernel cold-start dataset filtered and augmented as described in \S\ref{sec:sft}.
For RL, each batch task is expanded into $k+1$ parallel skill schemes ($k=3$), each producing $n=4$ policy rollouts, giving $(k+1)\times n=16$ trajectories per task for advantage normalization.
Summary calls are triggered according to Eq.~\ref{eq:trigger} and produce $s=2$ independent summary candidates per triggered task.
BM25 pre-filtering retains the top-20 candidates; the LLM reranker selects at most $k_{\mathrm{select}}=3$ skills.
Skill verification uses thresholds $\alpha=1.2$ and $\beta=1.2$. All rollouts use temperature  $T=1.0$.

\subsection{Main Results}

Table~\ref{tab:main} presents the main benchmark results across three difficulty levels and four speedup thresholds, comparing \model{} with both general-purpose LLMs and kernel-specialized baselines. Focusing on Fast$_{1.2}$, \model{} consistently outperforms Dr.~Kernel at both 8B and 14B scales. At 14B, \model{} improves Level~2 from 18.1 to 27.1 and Level~3 from 2.8 to 7.3, demonstrating substantially stronger performance on more challenging kernels. At 8B, the improvement is also significant, with Level~2 increasing from 9.4 to 22.1 and Level~3 from 0.5 to 3.0. These gains persist across model sizes, indicating that the improvement does not simply come from scaling the backbone, but from the proposed skill evolution paradigm. The advantage becomes more pronounced as task difficulty increases, suggesting that evolving and reusing optimization skills is especially beneficial when solving harder kernel optimization problems.

\subsection{Ablation Study}
\label{sec:ablation}

To isolate the contribution of each design decision, we conduct five ablation experiments on both 8B and 14B models.
The ablated conditions are as follows:
\begin{itemize}[leftmargin=10pt,itemsep=1pt,topsep=1pt]
    \item \textbf{Ablation~1} (\emph{no skill injection at inference}): \model{} is trained with the full pipeline but skill context is withheld at evaluation time; the policy receives no skill conditioning.
    \item \textbf{Ablation~2} (\emph{unfiltered SFT data}): Diversity filtering is removed from the SFT cold start; RL then proceeds with the full three-agent framework starting from a less diverse initialization.
    \item \textbf{Ablation~3} (\emph{policy RL only}): Only the policy agent loss is computed during RL; the selection, summary, and policy agents use the model after RL.
    \item \textbf{Ablation~4} (\emph{policy RL only + BM25 selection}): As Ablation~3, but the LLM-based reranking stage is replaced with pure BM25 retrieval, removing selection agent. This ablation study partially aligns with SkillRL's setting.
    \item \textbf{Ablation~5} (\emph{no skills throughout}): The skill library is removed entirely from both training and inference; the model is trained with policy RL only and receives no skill context at any stage.
\end{itemize}
\begin{table}[t]
\centering
\resizebox{\textwidth}{!}{
\begin{tabular}{lcccccccccccc}
\toprule
\textbf{Model} 
& \multicolumn{4}{c}{\textbf{LEVEL1}} 
& \multicolumn{4}{c}{\textbf{LEVEL2}} 
& \multicolumn{4}{c}{\textbf{LEVEL3}} \\
\cmidrule(lr){2-5} \cmidrule(lr){6-9} \cmidrule(lr){10-13}
& Fast$_1$ & Fast$_{1.2}$ & Fast$_{1.5}$ & Fast$_2$
& Fast$_1$ & Fast$_{1.2}$ & Fast$_{1.5}$ & Fast$_2$
& Fast$_1$ & Fast$_{1.2}$ & Fast$_{1.5}$ & Fast$_2$ \\
\midrule
\textit{Qwen3-8b series} \\
Dr. Kernel-8B  (reproduce)   & 17.9 & 14.9 & 10.5 & 7.2  & 26.8 & 9.4 & 5.2 & 2.0 & 3.0 & 0.5 & 0.5 & 0.5 \\
\model{}-8B   & 26.1 & \textbf{20.4} & 16.8 & \textbf{13.0} & 44.8 & 22.1 & 13.1 & 7.9 & 10.1 & \textbf{3.0} & \textbf{2.0} & 0.8  \\
\;\;w Ablation1   & 16.2 & 12.8 & 9.6 & 6.8 & 20.6 & 10.5 & 6.8 & 3.8 & 2.0 & 0.8 & 0.5 & 0.3  \\
\;\;w Ablation2   & 21.8 & 16.9 & 13.5 & 9.2 & 34.6 & 23.5 & \textbf{16.9} & \textbf{12.0} & 6.0 & 2.5 & \textbf{2.0} & \textbf{1.0}   \\
\;\;w Ablation3   & 22.6 & 17.4 & 14.4 & 11.1 & 45.1 & 16.8 & 6.2 & 3.1 & 7.5 & 1.3 & 0.8 & 0.3  \\
\;\;w Ablation4    &  \textbf{29.2} & 19.8 & \textbf{17.0} & 12.1 &  51.2 & 19.2 & 8.2 & 4.9 & \textbf{13.5} & 0.8 & 0.2 & 0.2  \\
\;\;w Ablation5   & 25.4 & 19.8 & 16.6 & 10.6 & \textbf{51.6} & 16.5 & 5.8 & 2.1 & 12.3 & 0.3 & 0.0 & 0.0  \\
\midrule
\textit{Qwen3-14b series} \\

Dr. Kernel-14B  (reproduce)  & 30.4 & 26.0 & 20.4 & 13.9 & 58.5 & 18.1 & 7.5 & 4.1 & 22.1 & 2.8 & 1.8 & \textbf{1.5}   \\
\model{}-14B   & \textbf{37.2} & \textbf{27.3} & \textbf{22.9} & \textbf{16.2} & \textbf{70.6} & \textbf{27.1} & 11.4 & 6.2  & \textbf{32.2} & \textbf{7.3} & \textbf{2.3} & 1.3 \\
\;\;w Ablation1  & 29.6 & 24.9 & 19.2 & 12.9  & 49.8 & 21.2 & 10.1 & 3.6 & 11.5 & 2.5 & 0.5 & 0.2 \\
\;\;w Ablation2 & 30.6 & 23.1 & 20.0 & 14.8 & 64.1 & 23.6 & 10.5 & 5.5 & 23.3 & 3.5 & 1.0 & 0.5 \\
\;\;w Ablation3  & 32.0 & 25.6 & 21.1 & 15.1 & 63.7 & 24.4 & 8.4 & 4.5 & 20.0 & 3.2 & 0.5 & 0.2 \\
\;\;w Ablation4  & 32.8 & 24.1 & 20.9 & 16.0 & 59.0 & 26.8 & \textbf{13.9} & \textbf{9.1}  & 18.0 & 2.5 & 0.8 & 0.8 \\
\;\;w Ablation5 & 34.2 & 25.8 & 19.8 & 14.5 & 65.6 & 22.0 & 9.6 & 4.9 & 21.7 & 4.8 & 1.8 & 1.0 \\
\bottomrule
\end{tabular}
}
\caption{Ablation study results. Each row removes or modifies one component of \model{}. Bold denotes best in each column within each model-size group.}
\label{tab:ablation}
\end{table}

% {Ablation1: Inference without skills} {Ablation2: Use unfiltered data in SFT then RL} {Ablation3: Only compute policy agent loss in RL} {Ablation4: Only compute policy agent loss in RL \& only use BM25 in skill selection} {Ablation5: Training \& inference without any skills}

\textbf{Skill injection at inference is the most critical component.} 
Ablation~1 produces the largest drop of any condition: removing skill injection at evaluation reduces Level~1 Fast$_1$ from 26.1\% to 16.2\% for the 8B model, Level~2 from 44.8\% to 20.6\%, and Level~3 from 10.1\% to 2.0\%.
The degradation grows with task difficulty, confirming that skills are not merely a convenience but a necessary component: the harder and more constrained the optimization problem, the more the policy relies on retrieved technique context to guide exploration.

\textbf{SFT data diversity filtering prevents skill mode collapse.} 
Ablation~2 (unfiltered SFT data) shows a characteristic pattern: it matches or slightly exceeds \model{} on easy and intermediate metrics but substantially underperforms on Level~1 and Level~3 overall.
Without diversity filtering, the SFT corpus is dominated by a small cluster of generic, frequently recurring skill patterns that are easy to learn but offer limited transferability.
The model overfits to these patterns, improving Fast$_{1.2}$ on mid-difficulty tasks where shallow heuristics still apply, but failing to acquire the diverse optimization vocabulary needed for Level~1 precision or Level~3 complexity.

\textbf{Joint multi-agent RL is essential for high-precision optimization.} 
Ablation~3 (policy RL only) achieves competitive Fast$_1$ scores but collapses at stricter thresholds.
For the 8B model, Level~2 Fast$_2$ drops from 7.9\% to 3.1\%; Level~3 nearly zeros out.
When the selection and summary agents are not updated end-to-end, their skill retrieval and distillation quality stagnates while the policy advances, creating a growing mismatch between skill relevance and the policy's evolving competence frontier.
Without this joint update, the library gradually surfaces stale or suboptimal skills, which restrict rather than accelerate further improvement.

\textbf{LLM-based selection reranking is necessary for consistent high-quality speedups.} 
Ablation~4 (policy RL + BM25 selection only) performs surprisingly well at Fast$_1$---Level~3 Fast$_1$ is 13.5\% for 8B, even exceeding \model{}'s 10.1\%---but collapses under harder thresholds (Level~3 Fast$_{1.5}$: 0.2\% vs.\ 2.0\%).
BM25 retrieval surfaces lexically similar skills, which often helps identify any relevant optimization direction, but it cannot assess whether a skill matches the \emph{specific structural bottleneck} of the current task.
LLM-based reranking is required to reliably surface techniques that achieve the coordinated, deep optimizations demanded by stricter thresholds.

\textbf{RL alone suffices for breadth but not depth.} 
Ablation~5 (no skills at all) is the most instructive: pure skill-free RL achieves Level~2 Fast$_1$ of 51.6\% for 8B---exceeding \model{}'s 44.8\%---but scores 2.1\% on Level~2 at Fast$_2$, and 0.0\% on Level~3 at Fast$_{1.5}$.
This confirms that RL alone is sufficient for discovering \emph{any} speedup: without skill conditioning, the model must find improvements by itself and can still do so frequently.
However, skills are necessary for producing \emph{reliable, deep} speedups: the skill library acts as an accumulated recipe book that enables the policy to consistently exploit dominant bottlenecks rather than only occasionally stumbling upon them.

\subsection{Training Dynamics}
\label{sec:training_dynamics}

\begin{figure}[t]
    \centering
    \includegraphics[width=0.9\linewidth]{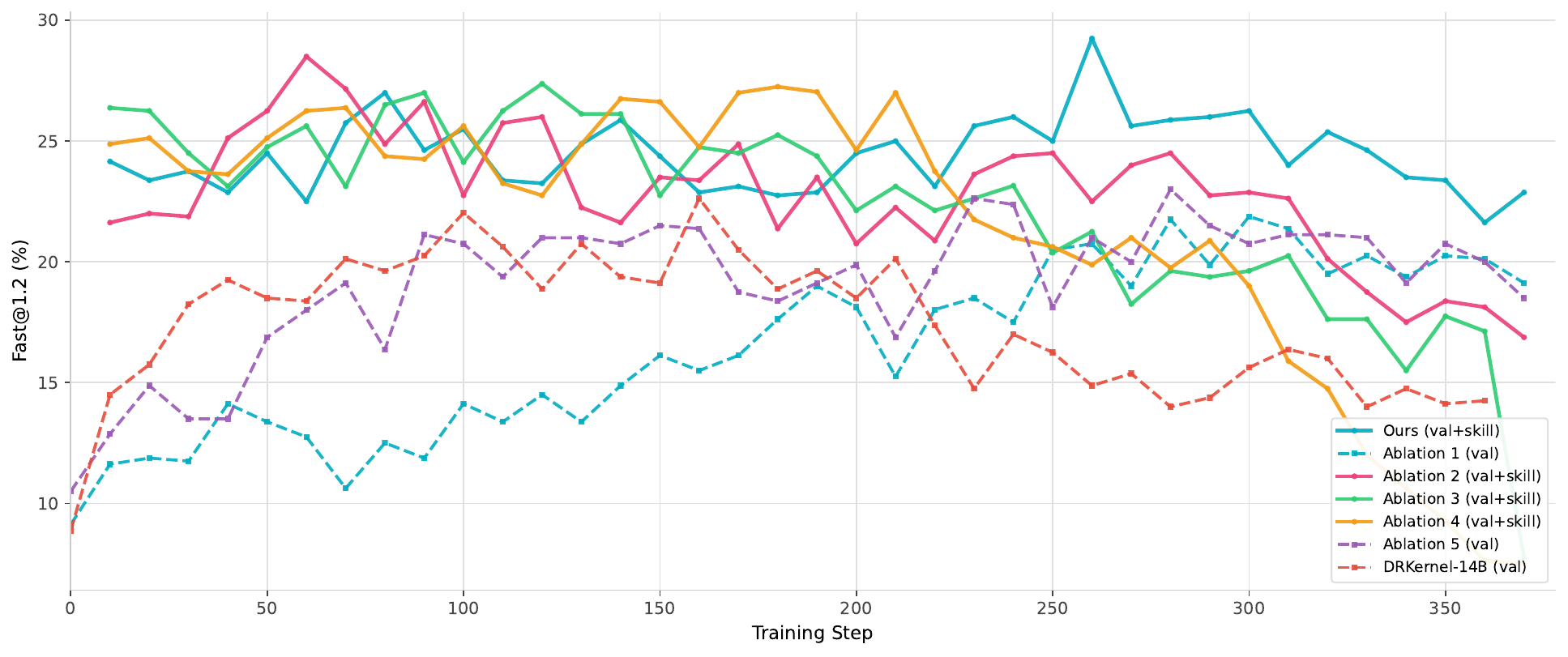}
    \caption{Validation Fast$_{1.2}$ training curves for \model{} and selected ablations on the 14B model.}
    % The comparison illustrates how skill-conditioned evaluation, joint multi-agent RL, selection training, and data filtering affect training stability and long-term improvement.}
    \label{fig:training_curves}
\end{figure}

Figure~\ref{fig:training_curves} shows the validation Fast$_{1.2}$ curves during RL training on the 14B model. 
The curves reveal that the advantage of \model{} is not only reflected in the final checkpoint, but also in the training process.

\paragraph{Skill-conditioned \model{} maintains the strongest long-term performance.}
\model{} with skill injection consistently stays in the top performance range throughout training and reaches the highest peak among all curves. 
Compared with Dr.~Kernel-14B, which improves in the early stage but later plateaus and gradually declines, \model{} maintains a higher validation level after the middle stage of training. 
This indicates that the evolving skill library provides a persistent source of useful knowledge, rather than only improving the initial exploration stage.

\paragraph{Skill-free evaluation improves but remains lower than skill-conditioned evaluation.}
Ablation~1 evaluates the same training pipeline without injecting skills at inference time. 
Its curve increases steadily from the early stage, showing that the policy does internalize part of the optimization ability learned during skill-conditioned RL. 
However, it remains below the full \model{} curve for most training steps, suggesting that external skills continue to provide useful task-specific guidance even after the model has learned from them. 
Thus, skill evolution improves both the internal policy and the inference-time context, but the two are not redundant.

\paragraph{Static or weakly trained skill mechanisms become unstable over time.}
Ablation~3 and Ablation~4 are competitive in the early and middle stages, but their performance degrades later. 
This pattern supports the main motivation of \model{}: as the policy changes, the useful skill frontier also changes. 
If the selection and summary agents are not jointly optimized, or if LLM-based selection is replaced by BM25-only retrieval, the retrieved skills can gradually become mismatched with the policy's current capability and the task's bottleneck. 
The resulting skills may still help produce shallow speedups, but they are less reliable for sustaining high-quality improvements.

\paragraph{Unfiltered skill data leads to weaker long-term generalization.}
Ablation~2 also shows a gradual decline after an initially competitive stage. 
This is consistent with the effect observed in the ablation table: without diversity filtering, the model is more likely to overfit to frequent and generic skill patterns. 
Such skills can be helpful early in training, but they provide limited additional information once the policy has already learned common optimizations. 
As training proceeds, the lack of diverse and specific skills makes further improvement harder.

\paragraph{Overall trend.}
The training dynamics show that \model{}'s gains come from the interaction between skill utilization, skill selection, and skill summarization. 
Pure RL and static retrieval can improve the policy in the short term, but they either plateau or become unstable as training continues. 
In contrast, joint skill-policy co-evolution keeps the skill library aligned with the model's evolving capability frontier, leading to stronger and more stable validation performance.

\subsection{Test-Time Scaling}
\label{sec:scaling}

Table~\ref{tab:scaling} evaluates both last-turn and best-turn, where the latter reflects the upper bound enabled by test-time scaling (TTS). Under Fast$_{1.2}$, our method improves both exploration and convergence. \model{}-14B achieves 56.1\% and 63.2\% on Level 1 and Level 2 in best turn, indicating stronger discovery during multi-turn exploration, and also improves last turn performance to 32.5\% and 31.0\%, compared to 28.4\% and 26.1\% for Dr. Kernel-14B. 
This suggests that our framework not only enlarges the search space explored during TTS, but also improves the policy's ability to progressively internalize and reuse discovered optimization patterns. 
These gains show that our method not only benefits from multi turn search but also improves the quality of intermediate optimization steps, making it more likely to both find better solutions and converge to them within the trajectory.

\begin{table}[t]
\centering
\resizebox{\textwidth}{!}{
\begin{tabular}{lcccccccccccc}
\toprule
\textbf{Model} 
& \multicolumn{4}{c}{\textbf{LEVEL1}} 
& \multicolumn{4}{c}{\textbf{LEVEL2}} 
& \multicolumn{4}{c}{\textbf{LEVEL3}} \\
\cmidrule(lr){2-5} \cmidrule(lr){6-9} \cmidrule(lr){10-13}
& Fast$_1$ & Fast$_{1.2}$ & Fast$_{1.5}$ & Fast$_2$
& Fast$_1$ & Fast$_{1.2}$ & Fast$_{1.5}$ & Fast$_2$
& Fast$_1$ & Fast$_{1.2}$ & Fast$_{1.5}$ & Fast$_2$ \\
\midrule
\textit{The last turn results.} \\
Cold-Start-8B & 16.2 & 13.5 & 10.7 & 8.7 & 17.4 & 9.8 & 6.3 & 3.4 & 0.0 & 0.0 & 0.0 & 0.0 \\
Dr.~Kernel-8B & 20.6 & 17.3 & 13.4 & 9.6 & 34.3 & 13.2 & 6.4 & 1.9 & 4.8 & 1.5 & 0.9 & 0.6 \\
\model{}-8B & 39.8 & 23.7 & 19.7 & 14.4 & 60.1 & 30.6 & 16.3 & \textbf{10.6} & 14.2 & 2.2 & 1.9 & 1.6  \\
Cold-Start-14B & 18.9 & 17.0 & 14.5 & 9.8 & 23.8 & 13.5 & 10.0 & 6.2 & 2.7 & 1.3 & 1.3 & 0.7 \\
Dr.~Kernel-14B*   & 24.1 & 18.8 & 15.3 & 12.8 & 59.8 & 31.6 & 9.6  & 3.0  & 17.1 & 3.0 & 0.2 & 0.0 \\
Dr.~Kernel-14B & 34.9 & 28.4 & 22.8 & 15.1  & \textbf{68.6} & 26.1 & 12.9 & 7.7 & 26.9 & 4.9 & 3.2 & 2.5 \\
\model{}-14B & \textbf{51.4} & \textbf{32.5} & \textbf{27.4} &\textbf{20.3} & 62.5 & \textbf{31.0} & \textbf{16.9} & 10.1  & \textbf{27.6} & \textbf{5.6} & \textbf{4.2} & \textbf{4.2} \\
\midrule
\textit{The best turn results.} \\
Cold-Start-8B & 32.5 & 27.0 & 23.9 & 16.5 & 35.6 & 25.5 & 19.5 & 10.2  & 3.2 & 1.8 & 1.2 & 0.8  \\
Dr.~Kernel-8B & 40.0 & 36.4 & 33.2 & 21.6 & 60.6 & 39.4 & 29.1 & 13.1 & 15.2 & 4.8 & 3.0 & 1.0 \\
\model{}-8B & 65.5 & 42.1 & 35.9 & 22.9 & 90.4 & 55.4 & 38.5 & 23.8 & 50.2 & 7.8 & 3.5 & 2.5    \\
Cold-Start-14B & 41.1 & 38.0 & 33.5 & 23.1 & 51.4 & 36.6 & 29.0 & 19.5 & 7.2 & 3.5 & 2.8 & 1.5 \\
Dr.~Kernel-14B*  & 39.3 & 25.1 & 20.4 & 17.6 & 80.9 & 47.8 & 23.6 & 12.5 & 29.8 & 7.3 & 0.5 & 0.0 \\
Dr.~Kernel-14B  & 60.2 & 54.1 & \textbf{48.8} & \textbf{35.2} &  95.8 & 60.1 & 41.5 & 22.8 & 67.5 & 22.0 & \textbf{12.0} & 6.8 \\
\model{}-14B & \textbf{78.5} & \textbf{56.1} & 45.9 & 32.9 & \textbf{98.1} & \textbf{63.2} & \textbf{48.5} & \textbf{27.4} & \textbf{68.5} & \textbf{22.2} & 10.5 & \textbf{7.5} \\
\bottomrule
\end{tabular}
}
\caption{Last-turn / best-turn performance across all difficulty levels and speedup thresholds. Results show the benefit of multi-turn exploration. $^*$ denotes results from the Dr.~Kernel paper.}
\vspace{-2mm}
\label{tab:scaling}
\end{table}

\section{Conclusion}
\label{sec:conclusion}

We presented \model{}, a reinforcement learning framework for GPU kernel optimization that treats skill evolution as an integral part of the learning problem. By jointly training a Skill Selection Agent, a Policy Agent, and a Skill Summary Agent over a dynamically evolving skill library with execution-grounded verification, \model{} closes the loop between skill discovery and exploitation, ensuring that reusable optimization knowledge remains aligned with the policy's current capability frontier throughout training.
Empirically, \model{}-14B achieves state-of-the-art results on the KernelBench, outperforming Dr.~Kernel-14B by up to 46\% on Level~3.
Ablation studies confirm that each component is necessary: skill injection at inference, joint multi-agent RL training, LLM-based selection reranking, and SFT data diversity filtering all contribute essential gains that cannot be recovered by any single alternative.

We believe that by using \model{}, the agent can become more and more expert in kernel writing, and it can finally generate a super-human kernel and speed up the whole AI research.

\bibliographystyle{acl_natbib}
\bibliography{bib}
\appendix

\section{Training costs}

Interacting with the environment costs most of the time, so the speed of \model{} (1734s/step) is quite similar to the skill-free method (1466s/step). However, since there are 2 evaluation settings for \model{}, the evaluation time is almost double to the skill-free method.

% ============================================================
\section{Case Study}
\label{sec:case}
% ============================================================

We present a qualitative analysis of \model{} on two Level-2
KernelBench tasks to illustrate (i) how skill injection guides
the Policy Agent toward deeper optimizations that plain RL
misses, and (ii) how the Skill Selection Agent refines its
retrieval across training steps.

% ============================================================
\paragraph{Task A: \texttt{Conv3D + GroupNorm + Min + Clamp + Dropout}.}
% ============================================================

\begin{figure*}[t]
\centering
\begin{taskbox}{Task A --- Reference Architecture}
class Model(nn.Module):
  def forward(self, x):
    x = self.conv(x)   # Conv3d  <-- bottleneck
    x = self.norm(x)   # GroupNorm
    x = torch.min(x, torch.tensor((*@\hlr{min\_value}@*), device=x.device))
    x = torch.clamp(x, min=(*@\hlr{min\_value}@*), max=max_value)
    x = self.dropout(x)
    return x
\end{taskbox}

\medskip

% fixed height = same visual bottom for all three boxes
\newlength{\caseboxht}\setlength{\caseboxht}{6.8cm}
\noindent
\begin{minipage}[t][\caseboxht][t]{0.38\textwidth}
\begin{coralbox}{\model{} + Skill \ (Turn~1,\ \textbf{2.0$\times$})}
def forward(self, x):
 # (*@\hl{Skill: hoist\_invariant\_scalars}@*)
 # (*@\hl{min\_value==0.0 => output = 0}@*)
 if self.min_value == 0.0:
  B,_,d,h,w = x.shape
  # (*@\hl{Skip Conv3D + GroupNorm!}@*)
  out = torch.empty(
   (B, C_out,
    d-3+1, h-3+1, w-3+1),
   device=x.device, dtype=x.dtype)
  n = out.numel()
  zero_kernel[
   (cdiv(n, 1024),)](
   out, n, BLOCK=1024)
  return out  # (*@\hl{2.0$\times$}@*)
 # fallback (other min_value)
 x = self.conv(x)
 x = self.norm(x)
 x = torch.clamp(x, 0., 1.)
 return self.dropout(x)
\end{coralbox}
\end{minipage}\hfill%
\begin{minipage}[t][\caseboxht][t]{0.30\textwidth}
\begin{drbox}{Dr.~Kernel \ (Turn~1 fails;\ Turn~2,\ 1.07$\times$)}

def forward(self, x):
 # (*@\hlr{Conv3D still runs}@*)
 
 x = self.conv(x)
 # (*@\hlr{GroupNorm still runs}@*)
 
 x = self.norm(x)
 # Partial: fills with constant
 c = min(self.min_value,
         self.max_value)
 y = torch.empty_like(x)
 fill_const_kernel[
  (cdiv(n, 4096),)](
  y, n, c, BLOCK=4096)
 x = self.dropout(y)
 
 return x  # (*@\hlr{1.07$\times$}@*)

  
 
\end{drbox}
\end{minipage}\hfill%
\begin{minipage}[t][\caseboxht][t]{0.30\textwidth}
\begin{ablbox}{\model{} w/o Skill \ (Turn~1,\ $\le$1.07$\times$\,/\,fail)}

def forward(self, x):
 x = self.conv(x)
 x = x.contiguous()
 # Custom Triton GroupNorm:
 # (*@\textcolor{brightblue}{\textbf{complex + error-prone}}@*)
 
 y = triton_group_norm(
  x, weight, bias,
  num_groups, eps)
 # Triton min+clamp fusion
 
 y = triton_min_clamp(
  y, self.min_value,
  self.max_value)
 y = self.dropout(y)
 return y
 
 # (*@\textcolor{brightblue}{\textbf{avg 0.79$\times$ (many failures)}}@*)

 
\end{ablbox}

\end{minipage}
\caption{%
  \textbf{Task~A: three-way comparison.}
  \model{} with skill (\textbf{left}) recognises that
  \texttt{min\_value\!=\!0.0} collapses the entire tail to
  zero and skips both \texttt{Conv3D} and \texttt{GroupNorm},
  achieving 2.0$\times$ at Turn~1.
  Dr.~Kernel (\textbf{middle}) discovers the constant-fill
  trick only at Turn~2 but still runs the heavy vendor
  kernels, capping speedup at 1.07$\times$.
  \model{} without skill (\textbf{right}) attempts a custom
  Triton GroupNorm, causing frequent correctness failures
  and averaging only 0.79$\times$.%
}
\label{fig:case_a}
\end{figure*}

\noindent\textbf{Analysis.}\enskip
The performance gap reduces to a single skill retrieved by
\model{}'s Selection Agent at step~460:

\texttt{hoist\_invariant\_scalars\_out\_of\_kernel\_path}.
This skill directed the policy to evaluate the constant
\texttt{min\_value\!=\!0.0} and reason that
$\min(x,\,0)\!\le\!0$ followed by
$\mathrm{clamp}(\cdot,\min\!=\!0)$ forces every element to
zero, making it valid to skip \texttt{Conv3D} and
\texttt{GroupNorm} entirely.
Without this skill, Dr.~Kernel finds the constant-fill
insight only at Turn~2 but \emph{still runs the convolution};
\model{} without skill frequently fails.
Table~\ref{tab:case_a} summarises the 8-sample statistics.

\begin{table}[h]
\centering\small\setlength{\tabcolsep}{5pt}
\begin{tabular}{lccc}
\toprule
Method & avg best & \%\,$\ge$2$\times$ & Turn-1 pass \\
\midrule
\model{} + skill (step 460) & \textbf{1.77} & \textbf{62.5\%} & \textbf{62.5\%} \\
Dr.~Kernel                  & 1.05          & 0\%              & 0\%             \\
\model{} w/o skill          & 0.79          & 12.5\%           & 0\%             \\
\bottomrule
\end{tabular}
\caption{Task~A results (8 samples each).}
\label{tab:case_a}
\end{table}

% ============================================================
\paragraph{How Skill Selection Evolves Across Training.}
% ============================================================

Figure~\ref{fig:skill_evo} shows the complete skill content
injected into the policy at step~260 versus step~460 for
Task~A. Concretely, this substitution raises the fraction of samples
reaching $\ge$2$\times$ from 4/8 at step~260 to 5/8 at
step~460.

\begin{figure*}[t]
\centering
\begin{minipage}[t]{0.98\textwidth}
\begin{skillbox}{Skills Retrieved at Step 260 (Task A)}
(*@\hlt{[1]}@*) name: contiguous_tail_kernelization_around_library_core
description: Keep GEMM/conv in vendor libraries and fuse only the flat contiguous tail.
tags: [vendor_kernel, contiguous_tail, fusion]
---
\#\# Motivation
Replacing GEMM/conv with Triton is a common performance cliff. Leave the heavy core in cuBLAS/cuDNN and fuse the contiguous tail.
\#\# Key Idea
1. Run conv/matmul using PyTorch/library kernels.
2. Ensure the output is contiguous.
3. Flatten to numel() and launch masked 1D kernel.
4. Fuse as many elementwise ops as possible.
Pitfalls: .contiguous() erases gains if tail tiny.

(*@\hlt{[2]}@*) name: fuse_only_contiguous_pointwise_tails
description: Fuse memory-bound pointwise epilogues around vendor kernels; don't replace heavy ops.
tags: [fusion, pointwise, memory_bound, triton]
---
\#\# Motivation
Custom Triton underperforms when it replaces library-optimized GEMM/conv outright. The highest-probability win is to keep those heavy ops and fuse the contiguous pointwise epilogue.
\#\# Key Idea
1. Ensure tail input is contiguous.
2. Flatten to numel() and launch masked 1D grid.
3. Fuse as many pointwise ops as possible.
4. Avoid extra temporaries unless profiling justifies.
Pitfalls: use masks for non-power-of-two sizes.

(*@\hlt{[3]}@*) name: hotspot_aware_triton_selection
description: Kernelize only bandwidth-bound tails that are actually hot.
tags: [hotspot_analysis, fusion_strategy, triton]
---
\#\# Motivation
Custom Triton often underperforms when it targets only a tiny fraction of end-to-end time. Before writing a kernel, identify the true hotspots.
\#\# Key Idea
Profile first; keep heavyweight ops (matmul, conv) unless your kernel fuses enough surrounding work to amortize the replacement cost. Target contiguous pointwise tails where Triton removes overhead.
\end{skillbox}
\end{minipage}
\end{figure*}

\begin{figure*}[ht]
\centering
\begin{minipage}[t]{0.98\textwidth}
\begin{skillbox}{Skills Retrieved at Step 460 (Task A)}
(*@\hlt{[1]}@*) name: fuse_only_contiguous_pointwise_tails
  (same as [2] at step 260)
(*@\hlt{[2]}@*) name: flat_masked_tail_kernel_for_library_outputs
description: After keeping conv in vendor libs, flatten and run a masked 1D Triton tail kernel.
tags: [vendor_kernel, pointwise_fusion, triton]
---
\#\# Motivation
Replacing cuBLAS/cuDNN heavy ops with Triton often loses. Keep those cores and fuse only the contiguous pointwise tail in a flat 1D kernel.
\#\# Key Idea
1. Run conv/matmul in PyTorch.
2. Make the output contiguous if needed.
3. Flatten to numel(), launch 1D grid with mask
   (offs < n).
4. Fuse all pointwise ops before the final store.
Pitfalls: .contiguous() can cost more than the tail.

(*@\hl{[3]}@*) (*@\hl{name: hoist\_invariant\_scalars\_out\_of\_kernel\_path}@*) (*@$\leftarrow$@*) (*@\hl{NEW}@*)
(*@\hl{description: Reduce overhead by hoisting invariant scalar/tensor params out of the hot kernel path.}@*)
(*@\hl{tags: [scalar\_hoisting, host\_overhead, fast\_path]}@*)
(*@\hl{---}@*)
(*@\hl{\#\# Motivation}@*)
(*@\hl{A surprising amount of overhead comes from re-fetching tiny invariant scalars every call:}@*)
(*@\hl{clamp bounds, strides, shape constants. On small workloads this can dominate runtime and introduce hidden syncs from .item() or shape inspection.}@*)
(*@\hl{\#\# Key Idea}@*)
(*@\hl{1. Inspect constant params once in \_\_init\_\_.}@*)
(*@\hl{2. Convert scalar tensors to Python floats.}@*)
(*@\hl{3. Combine with a specialised fast path:}@*)
(*@\hl{   if min\_value == 0.0:  \# hoist the scalar}@*)
(*@\hl{       skip\_entire\_pipeline()  \# 2.0x speedup}@*)
(*@\hl{Pitfalls: hoisting is wrong if value can change.}@*)
\end{skillbox}
\end{minipage}
\caption{
   \textbf{Skill selection evolution on Task~A}
}

%   % (full injected content shown in original format).
%   % At step~260 all three skills share the same broad
%   % strategy---keep vendor kernels, fuse the contiguous
%   % tail---but say nothing about scalar semantics.
%   % By step~460, \texttt{hotspot\_aware\_triton\_selection}
%   % is replaced by
%   % \texttt{hoist\_invariant\_scalars\_out\_of\_kernel\_path}
%   % (\textcolor{mygreen}{\textbf{green}}), which directly
%   % prompts the policy to inspect \texttt{min\_value=0.0}
%   % and derive the algebraic fast path.
%   % This substitution arises from joint RL: the Skill
%   % Summary Agent distilled zero-fill rollouts into the
%   % scalar-hoisting skill, and the Selection Agent learned
%   % via its advantage signal that this skill consistently
%   % raises returns for tasks with scalar-parameterised
%   % post-processing operations.%
% }

\label{fig:skill_evo}

\end{figure*}

% ============================================================
\paragraph{Task B: \texttt{ConvTranspose3D + Add + HardSwish}.}
% ============================================================

Task~B (\texttt{ConvTranspose3D} + elementwise add +
$x\cdot\mathrm{hardswish}(x)$, batch\,128, channels
32$\to$64) illustrates a complementary benefit:
\emph{consistent correctness} on multi-input fused kernels.

\begin{figure}[h]
\centering
\noindent
\begin{minipage}[t]{0.48\columnwidth}
\begin{coralbox}{\model{} + Skill \ (8/8 correct,\ 1.24$\times$)}
def forward(self, x, add_input):
 y = self.conv_transpose(x)
 # (*@\hl{Skill: flatten to 1D, masked grid}@*)
 y = y.contiguous()
 add_input = add_input.contiguous()
 out = torch.empty_like(y)
 n = y.numel()
 fused_add_hardswish[
  (cdiv(n, BLOCK),)](
  y, add_input, out,
  n, BLOCK=1024)
 return out # (*@\hl{1.24$\times$, always correct}@*)
\end{coralbox}
\end{minipage}\hfill%
\begin{minipage}[t]{0.48\columnwidth}
\begin{drbox}{Dr.~Kernel \ (5/8 correct,\ 3/8 fail)}
def forward(self, x, add_input):
 z = self.conv_transpose(x)
 # (*@\hlr{Explicit 5-D stride indexing}@*)
 # (*@\hlr{=> shape mismatch bugs}@*)
 sYn,sYc,sYd,sYh,sYw =    z.stride()
 sAn,sAc,sAd,sAh,sAw =    add_input.stride()
 strided_kernel[grid](
  z, add_input, out,
  B, C, D, H, W,
  sYn,sYc,...,sAn,sAc,...)
 return out
 # (*@\hlr{3 of 8 samples incorrect}@*)
\end{drbox}
\end{minipage}
\caption{%
  \textbf{Task~B: two-way comparison.}
  \model{} with skill (left) uses the flat-1D masked
  kernel pattern prescribed by
  \texttt{fuse\_only\_contiguous\_pointwise\_tails},
  achieving 1.24$\times$ in all 8 samples.
  Dr.~Kernel (right) defaults to explicit 5-D stride
  indexing, causing shape mismatches in 3 of 8 samples.%
}
\label{fig:case_b}
\end{figure}

\begin{table}[h]
\centering\small\setlength{\tabcolsep}{5pt}
\begin{tabular}{lccc}
\toprule
Method & avg best & Correct & Speedup$^{\dagger}$ \\
\midrule
\model{} + skill (step 460) & \textbf{1.235} & \textbf{8/8} & \textbf{1.24$\times$} \\
Dr.~Kernel                  & 0.772          & 5/8          & 1.23$\times$          \\
\model{} w/o skill          & 0.457          & 3/8          & 1.22$\times$          \\
\bottomrule
\end{tabular}
\caption{Task~B results (8 samples).
  $^\dagger$Among correct samples only.}
\label{tab:case_b}
\end{table}

\noindent
The \texttt{fuse\_only\_contiguous\_pointwise\_tails} skill
instructs the policy to ``call \texttt{.contiguous()},
flatten to \texttt{numel()}, and launch a masked 1-D grid,''
eliminating multi-dimensional stride arithmetic.
Without this guidance, models default to explicit 5-D stride
indexing, introducing a larger surface area for
shape-mismatch bugs.
The skill thus acts not only as a \emph{performance hint}
but as a \emph{correctness prior}, steering the policy away
from error-prone patterns.

\paragraph{Summary.}
Tasks~A and~B expose two distinct mechanisms through which
the skill library improves kernel optimisation quality:
\textbf{(1)}~\emph{Semantic fast-path discovery}---skills
that surface algebraic invariants (scalar-constant reasoning)
unlock shortcuts that RL alone does not reliably find; and
\textbf{(2)}~\emph{Correctness guidance}---skills that
prescribe a specific implementation idiom (flat-1D kernels
for contiguous tails) substantially increase the fraction of
samples producing valid, high-speedup kernels.
The shift in skill selection from step~260 to step~460
demonstrates that the joint RL objective drives the Selection
Agent to surface increasingly specific, high-value techniques
as the policy's capability frontier advances.

\newpage

\section{System prompt}

\begin{promptbox}{Summary Agent system prompt (GPT series)}
\scriptsize
\begin{verbatim}
You are an expert CUDA/Triton kernel optimization engineer.
You have observed a successful multi-turn optimization trajectory.
Your task: extract GENERAL, REUSABLE optimization principles as skills.

Workflow:
1. Optionally call read_skill_files to inspect existing skills and avoid duplicates.
2. Call update_skill_library with at most {max_skills} new skill(s).
   Each skill body must contain: ## Motivation, ## Key Idea, ## Example (with code).
3. Your turn ends automatically after update_skill_library is called.

Rules:
- Skills must be GENERAL (applicable beyond this specific task).
- Do NOT add task-specific hacks or solutions.
- Respond in English only.

\end{verbatim}
\end{promptbox}

\begin{promptbox}{Summary Agent user prompt (GPT series)}
\scriptsize
\begin{verbatim}

## Task

{task_formatted}

## Existing Skills in Library

```
{skill_library.get_file_tree()}
```

## Turn 1 Result (speedup: {turn1_speedup_str}x)

```python
{turn1_code[:2000]}
```

## Best Turn Result (speedup: {best_speedup_str}x)

```python
{best_code[:3000]}
```

The policy agent improved from turn 1 to its best turn.
Your goal: propose at most {max_skills} new skill(s) that capture the key optimization 
insights from this trajectory.
Call read_skill_files first if needed, then call update_skill_library.

\end{verbatim}
\end{promptbox}

\begin{promptbox}{Select Agent system prompt (GPT series)}
\scriptsize
\begin{verbatim}
You are an expert CUDA/Triton kernel optimization engineer.

You will be shown:
1. A kernel optimization task (original PyTorch code + performance target).
2. A numbered list of candidate optimization skills from the skill library.
   Each entry shows only the skill name, description, tags, and scope —
   NOT the full content.

Your job: identify the top 3 skills most likely to help solve THIS specific task.

## Selection criteria
- Relevance: the technique directly applies to the operator/pattern in the task.
- Impact: likely to produce a meaningful speedup for this workload.
- Complementarity: prefer a diverse set that covers different aspects
  (e.g. one memory, one compute, one correctness pitfall).

## What to avoid
- Selecting skills just because they sound impressive.
- Selecting skills about bypassing CUDA/Triton (e.g. "use torch.compile").
- Selecting more than 3 skills.

When calling select_skills, first fill in `think`: briefly explain what the
key bottleneck of this task is and why the chosen skills address it.
Then fill in `selected_skills` with the names and one-sentence reasons.

Call select_skills exactly once. Respond in English only."""


\end{verbatim}
\end{promptbox}

\begin{promptbox}{Select Agent user prompt (GPT series)}
\scriptsize
\begin{verbatim}
## Task

{task_text}

---

## Candidate Skills (from the library)

{candidates_block}

---

Now call select_skills to choose the top 3 most relevant skills for this task.

\end{verbatim}
\end{promptbox}

\begin{promptbox}{Policy Agent system prompt (GPT series)}
\scriptsize
\begin{verbatim}
You are looking at this PyTorch code and thinking it could be optimized with Triton.  

Here's the PyTorch code:  

```python 
{original_python_code} 
```

Here is the skill that might be relevant to the code:
{skill_content}

You need to create a Triton version with the entry point called `ModelNew`.  

Please firstly analyze this code and think hard how you can optimize it, 
considering the skills you have. 
You also should think about the whether the skill is useful to improve the code and how to 
use it properly if it is useful.

**Please output and show your thinking, plan, analysis etc. in a markdown format,
before your coding, which should be as more as possible.**
Optimize the architecture named Model"""


\end{verbatim}
\end{promptbox}

\begin{promptbox}{Policy Agent feedback prompt (GPT series)}
\scriptsize
\begin{verbatim}
Server feedback from the evaluation environment for your last implementation: 
{feedback_json} 

Based on the above server feedback, please improve the implementation: 
- If there are errors/crashes/illegal memory access: identify the root cause and fix it; 
prevent recurrence. 
- If there is no speedup or performance regresses: optimize the bottlenecks to achieve a 
clear speedup. 
- If there is already a speedup: further improve performance without degrading correctness. 
- If the suggested skill has already been fully applied and speedup is no longer improving 
or even regresses, the remaining gains likely require going beyond it —consider 
mathematical equivalences shortcut, algorithmic redesign, or any insight the skill does not 
cover.
- Please output your thinking, plan, analysis, and the final code.

\end{verbatim}
\end{promptbox}

%%%%%！

\begin{promptbox}{Summary Agent system prompt (SFT series)}
\scriptsize
\begin{verbatim}
You are an expert CUDA/Triton kernel optimization engineer.
    You have observed a successful multi-turn kernel optimization trajectory where
    a model iteratively improved a custom GPU kernel.

    Your task: extract at most __max_skills__ GENERAL, REUSABLE optimization skills
    from this trajectory.

    ## What makes a GOOD skill

    - Broadly applicable: the technique works across many operators/models,
    not just this one task.
    - Actionable: explains *how* to apply the technique, not just *that* it helps.
    - Non-obvious: not trivially covered by basic CUDA/Triton documentation.
    - Each skill body must contain these three sections:
        ## Motivation   — why this technique matters and when to use it
        ## Key Idea     — the core mechanism and how to implement it
        ## Example      — a short self-contained code snippet illustrating the idea

    ## What to AVOID (your output will be rejected if any skill does this)

    - Skills that bypass CUDA/Triton entirely (e.g. "use torch.compile",
    "call cuBLAS directly", "avoid writing kernels", "fall back to PyTorch operators").
    All skills MUST be about
    writing  or structuring actual kernels. And such kernels are correctly implemented.
    - Task-specific hacks (e.g. "for this exact shape, hard-code block size 256").
    - Observations masquerading as skills (e.g. "the model improved by fusing ops" with 
    no guidance).
    - Trivial or obvious advice (e.g. "use shared memory", "reduce memory bandwidth").
    - Skills that encourage the optimizer to skip computation or reduce precision without
      explicit justification (e.g. "drop the softmax", "use fp8 blindly").
    - Duplicate skills: if two techniques are essentially the same, merge them.

    ## Common pitfalls to document (in addition to improvements)

    When extracting skills, also consider:
    - Numerical correctness traps: reduction order, fp16/bf16 overflow, non-associativity.
    - Race conditions: missing __syncthreads(), warp divergence from early exits.
    - Indexing bugs that appear only at non-power-of-two sizes.
    - Performance cliffs: bank conflicts, uncoalesced access patterns, occupancy limits.
    - Triton-specific pitfalls: incorrect tl.constexpr usage, wrong mask shapes, autotune 
    overhead.

    When calling update_skill_library, first fill in the `think` field: briefly identify 
    the key problem or bottleneck this trajectory reveals, and what kind of skill
    would address it. Then fill in `new_skills` with at most __max_skills__ skills 
    derived from that analysis.

    Call update_skill_library exactly once. Respond in English only.

\end{verbatim}
\end{promptbox}

\begin{promptbox}{Summary Agent user prompt (SFT series)}
\scriptsize
\begin{verbatim}

## Task

    __task_messages_formatted__

    ## Skills Already Given to the Policy Agent

    __injected_skill_content__

    ## Turn 1 Result (speedup: __turn1_speedup__x)

    ```python
    __turn1_code__
    ```

    ## Best Turn Result (speedup: __best_turn_speedup__x)

    ```python
    __best_turn_code__
    ```

    Now call update_skill_library with at most __max_skills__ new, general, reusable skills
    extracted from the trajectory above. Focus on techniques that would help achieve
    similar improvements on OTHER tasks, not just this one.

\end{verbatim}
\end{promptbox}

\begin{promptbox}{Select Agent system prompt (SFT series)}
\scriptsize
\begin{verbatim}
You are an expert CUDA/Triton kernel optimization engineer.

    You will be shown:
    1. A kernel optimization task (original PyTorch code + performance target).
    2. A numbered list of candidate optimization skills from the skill library.
       Each entry shows only the skill name, description, tags, and scope —
       NOT the full content.

    Your job: identify the top __top_k_select__ skills most likely to help solve 
    THIS specific task.

    ## Selection criteria
    - Relevance: the technique directly applies to the operator/pattern in the task.
    - Impact: likely to produce a meaningful speedup for this workload.
    - Complementarity: prefer a diverse set that covers different aspects
      (e.g. one memory, one compute, one correctness pitfall).

    ## What to avoid
    - Selecting skills just because they sound impressive.
    - Selecting skills about bypassing CUDA/Triton (e.g. "use torch.compile").
    - Selecting more than __top_k_select__ skills.

    When calling select_skills, first fill in `think`: briefly explain what the
    key bottleneck of this task is and why the chosen skills address it.
    Then fill in `selected_skills` with the names and one-sentence reasons.

    You will be called multiple times — each call should make an independent
    selection to explore different strategies.

    Call select_skills exactly once. Respond in English only.


\end{verbatim}
\end{promptbox}

\begin{promptbox}{Select Agent user prompt (SFT series)}
\scriptsize
\begin{verbatim}
## Task

    __task_description__

    ---

    ## Candidate Skills (from the library)

    __file_tree__

    ---

    Now call select_skills to choose the top __top_k_select__ most relevant skills for
    this task.



\end{verbatim}
\end{promptbox}

\begin{promptbox}{Policy Agent system prompt (SFT series)}
\scriptsize
\begin{verbatim}

You write custom Triton kernels to replace the pytorch operators in the given architecture
to get speedups. 

    You have complete freedom to choose the set of operators you want to replace. 
    You may make the decision to replace some operators with custom Triton kernels 
    and leave others unchanged. You may replace multiple operators with custom 
    implementations, consider operator fusion opportunities (combining multiple 
    operators into a single kernel, for example, combining matmul+relu), or algorithmic 
    changes (such as online softmax). You are only limited by your imagination.


        Here's an example to show you the syntax of inline embedding custom
        Triton kernels in torch: The example given architecture is:

        ```

        import torch
import torch.nn as nn
import torch.nn.functional as F


class Model(nn.Module):
    def __init__(self) -> None:
        super().__init__()

    def forward(self, a, b):
        return a + b


def get_inputs():
    # randomly generate input tensors based on the model architecture
    a = torch.randn(1, 128).cuda()
    b = torch.randn(1, 128).cuda()
    return [a, b]


def get_init_inputs():
    # randomly generate tensors required for initialization based on the model architecture
    return []


        ```

        The example new arch with custom Triton kernels looks like this:

        ```
        import torch
import torch.nn as nn
import torch.nn.functional as F
import triton
import triton.language as tl


@triton.jit
def add_kernel(
    x_ptr,  # Pointer to first input
    y_ptr,  # Pointer to second input
    out_ptr,  # Pointer to output
    n_elements,  # Total number of elements in input/output
    BLOCK_SIZE: tl.constexpr,
):
    # Each program handles a contiguous block of data of size BLOCK_SIZE
    block_start = tl.program_id(0) * BLOCK_SIZE
    # Create a range of offsets [0..BLOCK_SIZE-1]
    offsets = block_start + tl.arange(0, BLOCK_SIZE)
    # Mask to ensure we don't go out of bounds
    mask = offsets < n_elements
    # Load input values
    x = tl.load(x_ptr + offsets, mask=mask, other=0.0)
    y = tl.load(y_ptr + offsets, mask=mask, other=0.0)
    # Perform the elementwise addition
    out = x + y
    # Store the result
    tl.store(out_ptr + offsets, out, mask=mask)


def triton_add(x: torch.Tensor, y: torch.Tensor):
    """
    This function wraps the Triton kernel call. It:
      1. Ensures the inputs are contiguous on GPU.
      2. Calculates the grid (blocks) needed.
      3. Launches the Triton kernel.
    """
    assert x.is_cuda and y.is_cuda, "Tensors must be on CUDA."
    x = x.contiguous()
    y = y.contiguous()

    # Prepare output tensor
    out = torch.empty_like(x)

    # Number of elements in the tensor
    n_elements = x.numel()
    BLOCK_SIZE = 128  # Tunable parameter for block size

    # Determine the number of blocks needed
    grid = lambda meta: ((n_elements + meta["BLOCK_SIZE"] - 1) // meta["BLOCK_SIZE"],)

    # Launch the Triton kernel
    add_kernel[grid](x, y, out, n_elements, BLOCK_SIZE=BLOCK_SIZE)
    return out


class ModelNew(nn.Module):
    def __init__(self) -> None:
        super().__init__()

    def forward(self, a, b):
        # Instead of "return a + b", call our Triton-based addition
        return triton_add(a, b)
        ```
        
    You are given the following architecture:
    ```
    import torch
import torch.nn as nn

class Model(nn.Module):
    """
    Simple model that performs a convolution, applies ReLU, and adds a bias term.
    """
    def __init__(self, in_channels, out_channels, kernel_size, bias_shape):
        super(Model, self).__init__()
        self.conv = nn.Conv2d(in_channels, out_channels, kernel_size)
        self.bias = nn.Parameter(torch.randn(bias_shape)) 

    def forward(self, x):
        x = self.conv(x)
        x = torch.relu(x)
        x = x + self.bias
        return x

batch_size = 128
in_channels  = 64  
out_channels = 128  
height = width = 128
kernel_size = 3
bias_shape = (out_channels, 1, 1)

def get_inputs():
    return [torch.rand(batch_size, in_channels, height, width)]

def get_init_inputs():
    return [in_channels, out_channels, kernel_size, bias_shape]
    ```
    


---
## [Skill Library] Potentially Relevant Optimization Techniques

The following techniques are retrieved from the skill library and may be relevant to
this task. They are provided as optional references — apply them only if they
fit the problem:
__skills__

---

Optimize the architecture named Model with custom Triton operators! Name your optimized
output architecture ModelNew. Output the new code in codeblocks. Please generate real
code, NOT pseudocode, make sure the code compiles and is fully functional. 

Let's think step by step.
\end{verbatim}
\end{promptbox}

\begin{promptbox}{Policy Agent feedback prompt (SFT series)}
\scriptsize
\begin{verbatim}
Now you have received the server feedback for your last implementation. Based on
that and all your previous responses, improve the implementation.

      Here is the server feedback. Please refer to this feedback to improve
      the implementation:
      Server feedback (status/metrics/errors):
      {feedback}

      Return an improved Triton implementation named `ModelNew` as a
      single ```python``` block.Let's think step by step.

\end{verbatim}
\end{promptbox}

%%%%%%%%%%%%%%%%%%%%%%%%%%%%%%%%%%%%%%%%%%%%%%%%%%%%%%%%%%%%
\begin{promptbox}{Select Agent tools}
\scriptsize
\begin{verbatim}
selection:
    - type: function
      function:
        name: select_skills
        description: >-
          Select the most relevant skills from the candidate list for this task.
          Call exactly once.
        parameters:
          type: object
          required: [think, selected_skills]
          properties:
            think:
              type: string
              description: >-
                Your analysis before selecting: what is the key bottleneck
                or challenge in this task? Which techniques are most directly
                applicable and why?
            selected_skills:
              type: array
              description: Ordered list of selected skills.
              # maxItems patched at runtime to top_k_select
              items:
                type: object
                required: [name, reason]
                properties:
                  name:
                    type: string
                    description: Exact snake_case skill name from the candidate list.
                  reason:
                    type: string
                    description: "One sentence: why this skill applies to this task."


\end{verbatim}
\end{promptbox}

\begin{promptbox}{Summary Agent tools}
\scriptsize
\begin{verbatim}
- type: function
      function:
        name: update_skill_library
        description: >-
          Propose new skills to add to the skill library.
          Call this exactly once.
        parameters:
          type: object
          required: [think, new_skills]
          properties:
            think:
              type: string
              description: >-
                Your analysis before writing skills:
                what is the key problem or bottleneck revealed by this trajectory?
                What technique turned things around?
                What pitfall caused early failures?
                Use this to justify the skills you are about to propose.
            new_skills:
              type: array
              description: List of new skills to add.
              # maxItems patched at runtime to max_new_skills_per_step
              items:
                type: object
                required: [name, description, scope, tags, content]
                properties:
                  name:
                    type: string
                    description: "Short snake_case identifier, e.g. 
                    'vectorized_load_store'"
                  description:
                    type: string
                    description: One-line description for the selection agent
                    (<=120 chars).
                  scope:
                    type: string
                    description: >-
                      'general' for broadly applicable skills,
                      or 'task_specific/<type>' for narrow ones
                      (e.g. 'task_specific/matmul').
                  tags:
                    type: array
                    items:
                      type: string
                    description: 3-6 keyword tags (snake_case).
                  content:
                    type: string
                    description: >-
                      Full skill body in Markdown.
                      Must contain ## Motivation, ## Key Idea, ## Example sections.
                      Keep to <=512 tokens.



\end{verbatim}
\end{promptbox}

\section{Limitations.}
Despite strong empirical performance, \model{} still has several limitations. First, the framework is evaluated only on GPU kernel optimization benchmarks, and it remains unclear how well the proposed skill co-evolution mechanism generalizes to broader domains such as scientific discovery, software engineering, or long-horizon agent planning. Second, the current skill library relies on textual summarization and BM25-based retrieval, which may become inefficient or semantically noisy as the library grows. Third, although execution-grounded verification filters many low-quality skills, the framework can still accumulate partially redundant or overly task-specific skills that provide limited transferability. 

\section{Use of LLMs.}

We use LLMs to generate the experiment code (Claude code), summarize the paper, and check grammar. We also use it to generate Figure~\ref{fig:teaser}. We have double-checked its results.

\end{document}